\documentclass{article}

\usepackage{amsmath,amsfonts,bm}

\def\1{\bm{1}}

\DeclareMathAlphabet{\mathsfit}{\encodingdefault}{\sfdefault}{m}{sl}
\SetMathAlphabet{\mathsfit}{bold}{\encodingdefault}{\sfdefault}{bx}{n}

\DeclareMathOperator*{\argmax}{arg\,max}

\newcommand{\trinorm}[1]{{\left\vert\kern-0.25ex\left\vert\kern-0.25ex\left\vert #1 
   \right\vert\kern-0.25ex\right\vert\kern-0.25ex\right\vert}}

\newcommand{\bc}{\boldsymbol{c}}

\newcommand{\bp}{\boldsymbol{p}}
\newcommand{\bq}{\boldsymbol{q}}

\newcommand{\bs}{\boldsymbol{s}}

\newcommand{\bv}{\boldsymbol{v}}

\newcommand{\bx}{\boldsymbol{x}}
\newcommand{\by}{\boldsymbol{y}}
\newcommand{\bz}{\boldsymbol{z}}

\newcommand{\bI}{\boldsymbol{I}}

\newcommand{\bV}{\boldsymbol{V}}
\newcommand{\bW}{\boldsymbol{W}}

\newcommand{\bbeta}{\bm{\beta}}

\newcommand{\cA}{\mathcal{A}}

\newcommand{\cC}{\mathcal{C}}

\newcommand{\cE}{\mathcal{E}}

\newcommand{\cH}{\mathcal{H}}

\newcommand{\cK}{\mathcal{K}}

\newcommand{\cM}{\mathcal{M}}

\newcommand{\cO}{\mathcal{O}}

\newcommand{\cT}{\mathcal{T}}

\newcommand{\bbB}{\mathbb{B}}

\newcommand{\bbN}{\mathbb{N}}

\newcommand{\bbR}{\mathbb{R}}

\newcommand{\bbZ}{\mathbb{Z}}
\newcommand{\bzero}{\mathbf{0}}

\newcommand{\pll}{\kern 0.56em/\kern -0.8em /\kern 0.56em}

\newcommand{\norm}[1]{\ensuremath{\left\| #1 \right\|}}

\newcommand{\supp}{\mbox{supp}}

    \usepackage[final]{neurips_2025}
\usepackage[utf8]{inputenc} % allow utf-8 input
\usepackage[T1]{fontenc}    % use 8-bit T1 fonts
\usepackage{titletoc}
\usepackage[pagebackref]{hyperref}   % hyperlinks
\usepackage{url}            % simple URL typesetting
\usepackage{booktabs}       % professional-quality tables
\usepackage{amsfonts}       % blackboard math symbols
\usepackage{nicefrac}       % compact symbols for 1/2, etc.
\usepackage{microtype}      % microtypography
\usepackage{xcolor}         % colors

\usepackage[ruled,vlined]{algorithm2e}

\hypersetup{colorlinks=true,linkcolor=red!70!black,linktocpage=false,citebordercolor=blue!70!black,citecolor=blue!70!black,anchorcolor=blue!70!black}

\usepackage{graphicx,subfig}
\usepackage{etoc}
\usepackage{amsmath}
\usepackage{amssymb}
\usepackage{mathtools}
\usepackage{amsthm}
\usepackage{multirow}
\theoremstyle{plain}
\newtheorem{theorem}{Theorem}[section]

\newtheorem{lemma}[theorem]{Lemma}
\newtheorem{corollary}[theorem]{Corollary}
\theoremstyle{definition}
\newtheorem{definition}[theorem]{Definition}

\newtheorem{example}[theorem]{Example}
\newtheorem*{main result}{Main Theorem}
\allowdisplaybreaks[4]

\usepackage{enumitem}
\definecolor{thistle}{rgb}{0.85, 0.75, 0.85}
\renewcommand{\geq}{\geqslant}
\renewcommand{\leq}{\leqslant}

\usepackage{wrapfig}
\usepackage{subfig}

\title{On the Expressive Power of Mixture-of-Experts for Structured Complex Tasks}

\author{
  Mingze Wang \\
  School of Mathematical Sciences,
  Peking University, Beijing, China \\
  \texttt{mingzewang@stu.pku.edu.cn} \\
  \And
  \hspace{-1.2cm} Weinan E \\
  \hspace{-1.2cm} Center for Machine Learning Research
  and School of Mathematical Sciences,
  Peking University, Beijing, China \\
  \hspace{-1.2cm} AI for Science Institute, Beijing, China \\
  \hspace{-1.2cm} \texttt{weinan@math.pku.edu.cn} \\
}

\begin{document}

\maketitle

\begin{abstract}
Mixture-of-experts networks (MoEs) have demonstrated remarkable efficiency in modern deep learning. Despite their empirical success, the theoretical foundations underlying their ability to model complex tasks remain poorly understood.
In this work, we conduct a systematic study of the expressive power of MoEs in modeling complex tasks with two common structural priors: low-dimensionality and sparsity.
For shallow MoEs, we prove that they can efficiently approximate functions supported on low-dimensional manifolds, overcoming the curse of dimensionality.
For deep MoEs, we show that $\cO(L)$-layer MoEs with $E$ experts per layer can approximate piecewise functions comprising $E^L$ pieces with compositional sparsity, i.e., they can exhibit an exponential number of structured tasks.
Our analysis reveals the roles of critical architectural components and hyperparameters in MoEs, including the gating mechanism, expert networks, the number of experts, and the number of layers, and offers natural suggestions for MoE variants.
\end{abstract}

% employing a gating network to sparsify and distribute diverse tasks among multiple experts.

% The abstract paragraph should be indented \nicefrac{1}{2}~inch (3~picas) on
% both the left- and right-hand margins. Use 10~point type, with a vertical
% spacing (leading) of 11~points.  The word \textbf{Abstract} must be centered,
% bold, and in point size 12. Two line spaces precede the abstract. The abstract
% must be limited to one paragraph.

\section{Introduction}

Mixture-of-experts (MoE) models~\citep{jacobs1991adaptive,jordan1994hierarchical} have recently achieved significant success in deep learning, particularly as a core architectural component of modern large language models (LLMs)~\citep{abdin2024phi,yang2024qwen2,liu2024deepseek,cai2025survey}.
These models have demonstrated strong capabilities across a wide range of complex and diverse tasks, including mathematical reasoning, logical inference, language understanding, and code generation. 
Despite their empirical success, the theoretical foundations underlying MoEs remain poorly understood, especially in their capacity to efficiently model complex tasks.

In both machine learning and applied mathematics,  it is widely recognized that although real-world tasks may appear complex, they often exhibit latent structures. Two prominent structural priors are: (1) {\em low-dimensional structure}: high-dimensional data typically lies on a manifold of much lower intrinsic dimension; (2) {\em sparse structure}: meaningful signals tend to admit sparse representations in suitable bases or dictionaries.
These structural priors have motivated numerous influential algorithms, including dimensionality reduction~\citep{tenenbaum2000global}, sparse regression via Lasso~\citep{tibshirani1996regression}, compressed sensing~\citep{donoho2006compressed}, and neural network pruning and compression techniques.

In this work, we investigate the expressive power of MoE networks for modeling complex tasks that exhibit either low-dimensional or sparse structure. {\bf Our contributions} are summarized as follows:

\begin{itemize}[leftmargin=2em]
\item {\bf Shallow MoE networks.} 
We prove that shallow MoE networks can efficiently approximate functions supported on {\em low-dimensional manifold}.
Theoretically, this task reduces to a collection of simpler approximation subproblems localized on low-dimensional subregions, along with an assignment problem that maps each input to the appropriate region.
We show that shallow MoE networks naturally implement this procedure, thereby avoiding the curse of dimensionality.
Our analysis reveal the complementary roles of the two core components in MoE: expert networks approximate localized subfunctions, while the gating mechanism ensures correct input-to-expert assignment.
Additionally, the analysis offers practical suggestions on MoE variants, such as the nonlinear gating, alternating MoE architectures with equivalent expressivity, and low-dimensional expert networks with auto-encoding.
	
\item {\bf Deep MoE networks.} 
We formalize complex tasks as piecewise functions, and focus on a broad class of structured tasks exhibiting \emph{compositional sparsity}: where each subtask depends on only a small subset of input coordinates, and the overall task is a hierarchial composition of these subtasks.
We demonstrates that a depth-$\cO(L)$ MoE network with $E$ expert per layers can efficiently approximate piecewise functions with $E^L$ distinct pieces, i.e, it can exhibit an exponential number of structured tasks. 
Moreover, our analysis elucidates the distinct roles of network depth $L$ (which enables hierarchical composition) and expert count $E$ (which enables subtask specialization).

\item {\bf Unified insights.} 
Our theoretical results reveal that MoE networks can effectively discover the underlying structure priors in the complex tasks (such as low-dimensionality or sparsity), and subsequently decompose them into simpler subproblems, each solved by specialized experts.

\end{itemize}

% Intuitively, MoEs can capture the low-dimensional or sparsity sturcture and  ``Divide and conquer'' the complex task to a series of simple problems.

\section{Related Works}

{\bf Theoretical understanding of MoE}.
% As previously noted, MoE networks are widely gained widespread adoption in modern deep learning, particularly within LLMs; however, their theoretical foundations remain comparatively underexplored.   
\citet{chen2022towards} analyzed the training dynamics of shallow MoE networks with softmax gating on clustered datasets, emphasizing the importance of expert nonlinearity and data structure.
~\citep{baykal2022theoretical} showed that sparsely activated networks can achieve approximation performance comparable to dense networks, and offered a computationally efficient alternative.
\citet{dikkala2023benefits} examined the impact of learnable routing mechanisms in MoEs, establishing their benefits.
\citet{li2024theory} investigated MoE in continual learning, using overparameterized linear regression to show their adaptability across tasks.
A comprehensive survey of recent theoretical advances is presented in~\citet{mu2025comprehensive}.
In contrast to these prior works, we focus on the expressive power of both shallow and deep MoE networks for broad classes of structured functions.

% Sparse and low-dimensional structures are fundamental principles in machine learning and applied mathematics, driving the development of numerous successful algorithms.

{\bf Low-dimensional structure.}
The {\em manifold hypothesis} posits that high-dimensional data in real world (e.g., images, speech, and text) typically lies on a manifold of much lower intrinsic dimensionality than the ambient space.
This perspective motivates various algorithmic approaches:
{\em (i) Dimensionality reduction} techniques~\citep{tenenbaum2000global,roweis2000nonlinear,belkin2003laplacian}, which aim to uncover and utilize such low-dimensional structures.
{\em (ii) Representation learning} methods like Autoencoders and Variational Autoencoders~\citep{hinton2006reducing,kingma2013auto}, which seek compact and informative representations aligned with low-dimensional manifold.

{\bf Sparse structure.} 
It is widely believed that meaningful signals often admit sparse representations in appropriate bases or dictionaries.
This principle underpins many influential algorithms, such as Lasso~\citep{tibshirani1996regression}, Compressed Sensing~\citep{donoho2006compressed,candes2008introduction}, and Sparse Coding~\citep{olshausen1996emergence,elad2006image}, which have been widely applied across domains.

From a theoretical standpoint, the prevalence of sparsity and low-dimensionality has inspired recent studies on the expressive power of deep networks under structural assumptions.
For example,~\citet{mhaskar2016deep,poggio2023deep} analyzed dense neural networks approximating functions with compositional sparsity, demonstrating how sparsity mitigates the curse of dimensionality~\citep{bellman1966dynamic,bach2017breaking}.
\citet{wang2024understanding} studied the expressivity of Transformer models~\citep{vaswani2017attention} for modeling long but sparse memories, showing the model's capacity to overcome the curse of memory.
\citet{shaham2018provable,chen2019efficient} examined the approximation power of dense networks for functions supported on low-dimensional manifolds.
In contrast to these works, we investigates the expressive power of MoE networks in approximating complex functions exhibiting either sparse or low-dimensional structure.

\section{Preliminaries}

\begin{wrapfigure}{r}{0.4\textwidth}
    
    \vspace{-.3cm}
    
    \centering
    \includegraphics[width=0.3\textwidth]{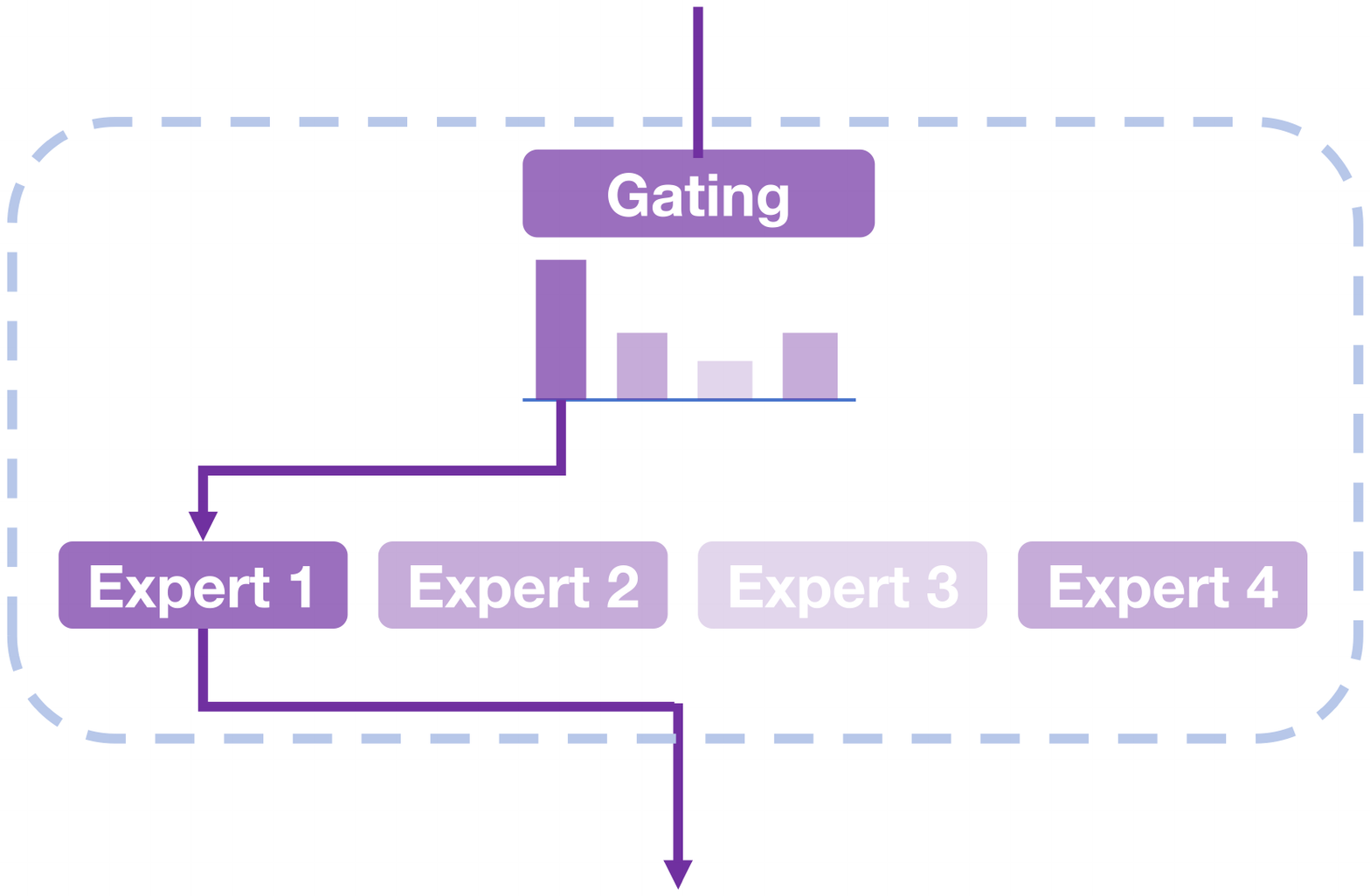}

     \vspace{-.3cm}
     
    \caption{\small Illustration of an MoE layer.}
    \label{fig: moe network}
    
    \vspace{-1.cm}
    
\end{wrapfigure}
\paragraph{Basic notations.} 
Let $f:\Omega\to\bbR$ be a continuous function defined on a compact set $\Omega$.
Its $L_{\infty}$ norm is defined as $\norm{f}_{L_{\infty}}:=\sup_{\bx\in\Omega}|f(\bx)|$.
We use standard asymptotic notations $\cO(\cdot),\Omega(\cdot),\Theta(\cdot)$ to hide the constants independent of the primary problem size (typically denoted by $m$), and the notations
$\tilde{\cO}(\cdot),\tilde{\Omega}(\cdot),\tilde{\Theta}(\cdot)$ further hide logarithmic factors. 
For a positive integer $n$, let $[n]=\{1,\cdots,n\}$.
For $a,b\in\bbR$, define $a\wedge b=\min\{a,b\}$ and $a\vee b=\max\{a,b\}$.

\subsection{MoE networks}

\paragraph{\bf MoE components.} An MoE layer consists of two primary components:
\begin{itemize}[leftmargin=2em]
    \item {\bf Expert networks:} A collection of $E$ expert networks, each implemented as a dense feedforward ReLU neural network: $f^{(1)},\cdots,f^{(E)}:\bbR^{d_\text{in}}\to\bbR^{d_\text{out}}.$ 
    % For an input $\bx\in\bbR^{d_\text{in}}$, the output of the $e$-th expert is denoted by $f^{(e)}(\bx)$.
    
    \item {\bf Gating network:} A gating function $g:\bbR^{d_\text{in}}\to\bbR^E$. In most existing MoE models~\citep{fedus2022switch,du2022glam,yang2024qwen2technicalreport}, $g$ is linear due to its simplicity and empirical effectiveness:
    $g(\bx)=\bW_R\bx$, where $\bW_R\in\bbR^{E\times d_\text{in}}$.
\end{itemize}

\paragraph{MoE operation.} 
Given an input $\bx\in \bbR^{d_{\text{in}}}$, an MoE layer performs the following operations, as illustrated in Figure~\ref{fig: moe network}:

\begin{itemize}[leftmargin=2em]
    \item {\bf Expert selection.} The gating network computes routing scores $g(\bx)\in\bbR^E$, and selects the top-$K$ experts with the highest scores:
    $$\cK:=\arg{\rm TopK}(g(\bx)),$$
    where $\arg{\rm TopK}(\bz)$ returns the indices of the $K$ largest entries of $\bz$.
    
    \item {\bf Expert computation and aggregation.} Each selected expert $k\in\cK$ computes its output $f^{(k)}(\bx)$. The final output is a weighted combination: $$\by=\sum_{k\in\cK}\alpha_k(\bx) f^{(k)}(\bx),$$
    where the weight are defined via $\alpha_k(\bx)=\frac{\exp(g_k(\bx))}{\sum_{j\in\cK}\exp(g_j(\bx))}$.
\end{itemize}

Notably, only $K$ expert networks are activated per input.
Without loss of generality, we focus throughout this paper on {\bf the case} $K=1$, as  the extension to arbitrary $K\leq E$ is straightforward.

{\bf Hypothesis class $\cH_{l,m}^{L,E}$}. 
We define $\cH_{l,m}^{L,E}$ as the class of depth-$L$ neural networks composed of stacked $L$ MoE layers:
\begin{equation}\label{equ: MoE class}
    h^{(L)}\circ h^{(L-1)}\circ\cdots\circ h^{(1)},
\end{equation}
where each $h^{(\ell)}$ is an MoE layer consisting of a linear gating network $g^{(\ell)}$ and $E$ expert networks $f^{(\ell,e)}$ ($e\in[E]$), each being an $l$-layer, $m$-width dense ReLU neural network.

\subsection{Classical approximation results}

{\bf Approximation error notation.}
Let $\cE_{l,m}^{\rm FFN}(f)$ denote the $L^\infty$ approximation error of a target function $f:\Omega\to\bbR$ using $l$-layer, $m$-width dense ReLU neural networks.

{\bf $\cC^K$ space.}
Let $D,K\in\bbN$, and $\Omega\subset\bbR^D$ be compact. The space $\cC^K(\Omega)$ consists of all functions $f$ such that
\begin{equation}
    \|f\|_{\cC^K(\Omega)}=\max\limits_{0\leq\|\bbeta\|_1\leq K}\|D^{\bbeta} f\|_{L_{\infty}(\Omega)}<\infty
\end{equation}
where $D^{\bbeta} f$ denotes the partial derivatives of order $\bbeta=(\beta_1,\cdots,\beta_D)\in\bbZ_+^D$.
The space of {\bf smooth functions} is defined as $\cC^\infty(\Omega)=\bigcap_{K\in\bbN}\cC^{K}(\Omega)$.
Additionally, the {\bf smoothness exponent} of $f:\Omega\to\bbR$ is defined as 
\begin{equation}\label{equ: smoothness, euclidean}
    \kappa(f):=\sup\{K\in\bbN:f\in\cC^K(\Omega)\}.
\end{equation}

The following result summarizes the classical approximation rate of two-layer ReLU networks for $\cC^K$ functions~\citep{mao2023rates,yang2024optimal}:

\begin{theorem}\label{thm: C^k approximation rate}
Let $D,K\in\bbN$, and $\Omega\subset\bbR^D$ be compact. 
For any $f\in\cC^K(\Omega)$ and $m\in\bbN$, there exits a two-layer ReLU neural network $f_m$ with $m$ hidden neurons such that
\begin{align*}
    \cE_{2,m}^{\rm FFN}(f)
    \leq\|f-f_m\|_{L_\infty(\Omega)}\leq 
    \begin{cases}
        \cO\big(m^{-\frac{K}{D}}\big),\quad \text{ if } K<\frac{D+3}{2},\\
        \tilde{\cO}\big(m^{-\frac{1}{2}}\big),\quad \text{ otherwise}.
    \end{cases}
\end{align*}
\end{theorem}

When the smoothness of the target function is relatively low, i.e., $K\ll D$, the approximation rate $\cO\big(m^{-\frac{K}{D}}\big)$ reveals that two-layer networks suffer from the curse of dimensionality (CoD). 
% This implies that two-layer networks suffer from the curse of dimensionality (CoD).
% Moreover, in this regime, it has been shown that for any approximation method, its minimax approximation rate is $\cO\big(m^{-\frac{K}{D}}\big)$, also suffering from CoD.

\section{Theory for Shallow MoE Networks}\label{section: shallow MoE}

In this section, we study the efficiency of shallow MoE networks in approximating functions supported on a low-dimensional manifold $\cM$.

\begin{wrapfigure}{r}{0.35\textwidth}
    \vspace{-1.cm}
    \centering
    \includegraphics[width=0.4\textwidth]{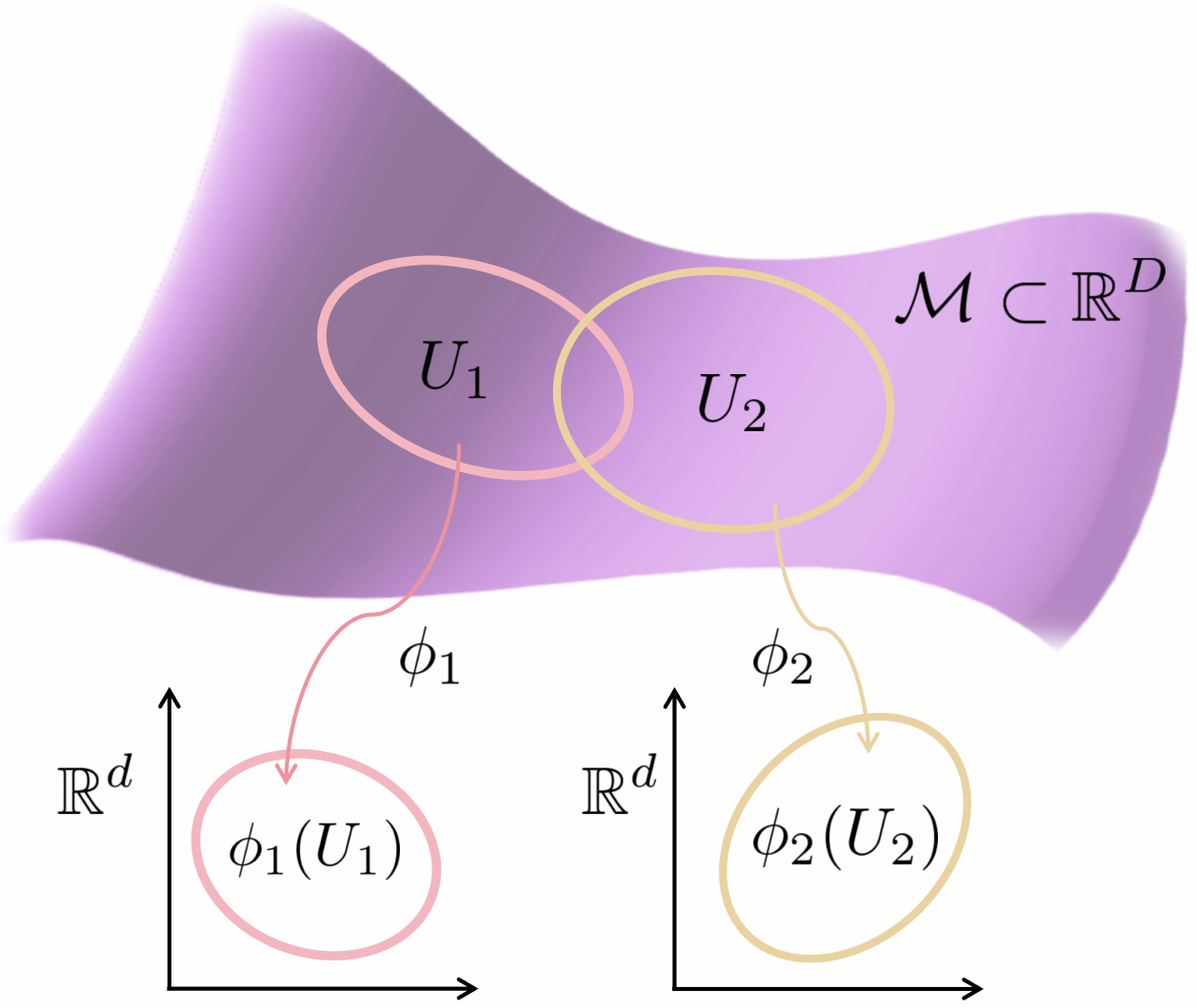}
    \vspace{-.6cm}
    \caption{\small A $d$-dimensional manifold $\cM$ in $\bbR^D$.}
    \vspace{-1.cm}
\end{wrapfigure}
\subsection{Manifold in Euclidean Space}

Let $\cM$ be a $d$-dimensional smooth manifold embedded in $\bbR^D$.
We begin by reviewing several standard definitions.

\begin{definition}[Chart and Atlas]\ 
    \begin{itemize}[leftmargin=2em]
        \item A {\em chart} for $\cM$ is a pair $(U,\phi)$ such that $U\subset\cM$ is open and $\phi:U\to\bbR^d$, where $\phi$ is a homeomorphism (i.e., bijective, $\phi$ and $\phi^{-1}$ are both continuous).
        $U$ is called a coordinate neighborhood, and $\phi$ is the associated coordinate map.
        \item An {\em atlas} of $\cM$ is a collection $\{(U_\alpha,\phi_\alpha)\}_{\alpha\in\cA}$ of charts such that $\cup_{\alpha\in\cA}U_\alpha=\cM$.
    \end{itemize}
    
\end{definition}

An atlas $\{(U_\alpha,\phi_\alpha)\}_{\alpha\in\cA}$ is called {\em smooth} if for any overlapping charts $(U_\alpha,\phi_\alpha)$ and $(U_{\alpha'},\phi_{\alpha'})$, the transition maps $\phi_\alpha\circ\phi_{\alpha'}^{-1}$ and $\phi_{\alpha'}\circ\phi_\alpha^{-1}$ are smooth functions.

\begin{definition}[Smooth manifold]
The manifold $\cM$ is called {\em smooth} if it has a smooth altas.
\end{definition}

We now introduce the partition of unity, which can divide the manifold into regular subregions.

\begin{definition}[Partition of unity]
    Let $\{U_\alpha\}_{\alpha \in \cA}$ be an open cover of $\cM$.
    A {\bf partition of unity} of $\cM$ w.r.t this cover is a family of nonnegative smooth functions $\rho_\alpha:\cM\to[0,1]$ for $\alpha\in\cA$ such that: 
    \begin{itemize}[leftmargin=2em]
        \item {\em (i)} for all $\alpha\in\cA$, $\rho_\alpha$ has compact support and $\supp(\rho_\alpha)\subset U_\alpha$;
        \item 
    {\em (ii)} for every $\bx\in\cM$, only finitely many $\rho_\alpha(\bx)$ are nonzero;
    \item 
    {\em (iii)} for all $\bx\in\cM$, $\sum_{\alpha\in\cA}\rho_\alpha(\bx)=1$.
    \end{itemize}
\end{definition}

\begin{theorem}[Existence of a partition of unity]\label{thm: partition of unity}
Let $\{U_\alpha\}_{\alpha\in\cA}$ be an open cover of a smooth manifold $\cM$. Then there exists a partition of unity $\{\rho_\alpha\}_{\alpha\in\cA}$ of $\cM$ w.r.t. $\{U_\alpha\}_{\alpha\in\cA}$.
\end{theorem}

We next define the smoothness of a function defined on a manifold.

\begin{definition}[Function on the manifold]
Let a function $f:\cM\to\bbR$, and $\{(U_\alpha, \phi_\alpha)\}_{\alpha \in \cA}$ be a smooth atlas of $\cM$. Its smoothness $\kappa(f)$ is defined by $\kappa(f):=\inf_{\alpha\in\cA}\kappa(f\circ\phi_{\alpha}^{-1})$, where the smoothness of each $f\circ\phi_{\alpha}^{-1}$ is defined as Equation~\eqref{equ: smoothness, euclidean}.
\end{definition}

In this paper, we focus on {\bf smooth compact manifolds}. Due to the compactness of $\cM$, its atlas consists of a finite collection of charts, denoted by $\{(U_i,\phi_i)\}_{i\in[E]}$. Additionally, we can let $\phi_i(U_i)\subset[0,1]^d$.
By Theorem~\ref{thm: partition of unity}, there exists a corresponding partition of unity $\{\rho_i\}_{i\in[E]}$.

Compact smooth manifolds admit an atlas with strong geometric regularity as below, which is detailed in Appendix~\ref{appendix: proofs for shallow MoE}.

\begin{example}[Highly regular atlas]\label{example: manifold without boundary}
Let $\cM$ be a compact smooth manifold. Then there exists a highly smooth atlas $\{(U_i,\phi_i)\}_{i\in[E]}$ such that each map $\phi_i: U_i\to[0,1]^d$ is a {\em linear function}. Thus, each $\phi_i$ satisfies $\kappa(\phi_i)=\infty$ (when viewed as a function in $\bbR^D$).
% Let $\cM$ is a smooth, compact manifold without boundary. Then there exists a smooth atlas $\{(U_i,\phi_i)\}_{i\in[E]}$ such that each map $\phi_i: U_i\to[0,1]^d$ is a {\em linear function}. Thus, each $\phi_i$ satisfies $\kappa(\phi_i)=\infty$ (when viewed as a function in $\bbR^D$).
\end{example}

Motivated by this example, we define a broad class of regular atlas:

\begin{definition}[Regular atlas]\label{def: regular manifold}
    A atlas $\{(U_i,\phi_i)\}_{i\in[E]}$ of compact manifold $\cM$ is called regular, if each map $\phi_i$ has the smoothness $\kappa(\phi_i)>\frac{D+3}{2}$.
    % (when viewed as a function in $[0,1]^d$).
\end{definition}

\subsection{Theoretical results and insights}

\begin{theorem}[Main result]
\label{thm: 1-layer, general}
    Let $\cM$ be a {\em compact}, $d$-dimensional smooth manifold in $\bbR^D$, with a regular atlas $\{(U_i, \phi_i)\}_{i\in [E]}$ (Definition~\ref{def: regular manifold}).
    Let the target function $f:\cM\to\bbR$.
    Then for any $m\geq \Omega\big(E^{2}\big)$,
    there exists a depth-$2$ MoE network $\Psi\in\cH_{3,m}^{2,E}$, with $E$ experts per layer, each being is a $3$-layer $m$-width dense networks,
    % \sum_{i\in[E]}\|\rho_i\|_{\cC^{\infty}(\cM)}^{2}\sum_{i\in[E]}\|\rho_i\|_{\cC^{\infty}(\cM)}^{2}
    such that:
    \begin{align*}
        &\|f-\Psi\|_{L_\infty(\cM)}
        \leq\max_{i\in[E]} \cE_{3,m}^{\rm FFN}\big(f|_{U_i}\big)
        \\\leq&
        \max_{i\in[E]}\underbrace{\cE_{2,m}^{\rm FFN}\big(f|_{U_i}\circ\phi_i^{-1}\big)}_{\text{approximate a {\color{red!70!black} low-dimensional} function}}+\max_{i\in[E]}\!\!\!\!  \underbrace{\cE_{2,m}^{\rm FFN}\big(\phi_i\big)}_{\text{approximate a {\color{red!70!black} high-order smooth} map}}\!\!\!\!\!\!\norm{f|_{U_i}\circ \phi_i^{-1}}_{\cC^1([0,1]^d)}
        \\\leq&
        \max_{i\in[E]}\ 
        \tilde{\cO}\Big(m^{-\frac{\color{red!70!black}\kappa(f|_{U_i})}{\color{red!70!black} d}\wedge\frac{1}{2}}\Big).
    \end{align*}
% Here, the term $\cE_{2,m}^{\rm FFN}\big(f|_{U_i}\circ\phi_i^{-1};[0,1]^r\big)$ admits the rate in Theorem~\ref{thm: C^k approximation rate} with $D=r$ and $K=\kappa(f|_{U_i})$; 
% the term $\cE_{2,m}^{\rm FFN}\big(\phi_i;U_i\big)$ admits the rate in Theorem \ref{thm: C^k approximation rate} with $D=d$ and $K>\frac{d+3}{2}$.
\end{theorem}

Theorem~\ref{thm: 1-layer, general} demonstrates that depth-2 MoE networks can efficiently approximate functions supported on low-dimensional manifolds.
The total approximation error decomposes into two components: (i) approximation of low-dimensional target functions $f|_{U_i}\circ\phi_i^{-1}$'s; (ii) approximation of smooth coordinate maps $\phi_i$'s.
Both subproblems are significantly simpler than approximating the original high-dimensional function directly, enabling MoE networks to overcome the curse of dimensionality.

\begin{table}[!ht]
    \caption{Comparison of approximation rates between shallow MoE and shallow dense networks. For MoE, $m$ is the width of each expert networks; for dense networks, $m$ is the width of the hidden layer.}
    \centering
    \begin{tabular}{c|c}
        \hline\hline
        Shallow MoE networks (Theorem~\ref{thm: 1-layer, general}) & Shallow dense networks (Theorem~\ref{thm: C^k approximation rate}) \\ \hline
        $\max\limits_{i\in[E]}\tilde{\cO}\Big(m^{-\frac{\color{red!70!black}\kappa(f|_{U_i})}{\color{red!70!black} d}\wedge\frac{1}{2}}\Big)$ & $\tilde{\cO}\left(m^{-\frac{\color{red!70!black} \kappa(f)}{\color{red!70!black} D}\wedge\frac{1}{2}}\right)$
        \\\hline\hline
    \end{tabular}
    \label{tab: approximation rate comparison}
\end{table}

{\bf Improved efficiency over dense networks.}
Table~\ref{tab: approximation rate comparison} highlights the superior approximation efficiency of MoEs. 
In the regime where the target $f$ has limited smoothness, i.e., $\kappa(f)\ll D$, dense networks suffer from the curse of dimensionality, as the approximation rate $\kappa(f)/D$ deteriorates with ambient dimension $D$.
In contrast, MoE networks achieve rates governed by the intrinsic dimension $d\ll D$ and local smoothness $\kappa(f|_{U_i}) \geq \kappa(f)$, thereby substantially improving approximation efficiency and achieving faster approximation rate: $\frac{\kappa(f|_{U_i}\circ\phi_i^{-1})}{d}\wedge\frac{1}{2}\gg\frac{\kappa(f)}{D}$.

{\bf Key insight.}
The proof of Theorem~\ref{thm: 1-layer, general} (deferred to Appendix~\ref{appendix: proofs for shallow MoE}) reveals several insights into the mechanisms of MoE networks:

\begin{center}
    \centering
    \em MoE networks achieve efficient approximation by decomposing a complex approximation problem 
    into multiple localized approximation subproblems, as well as a simple assignment task.
\end{center}

% MoE achieves efficient approximation by decomposing the problem into {\bf several simple localized approximation problems and a simple assignment problem}.
% By the Partition of Union, the mainifold $\cM$ is naturally divided into $E$ simple , which to $[0,1]^d$.

Recall that a depth-2 MoE (Eq.~\eqref{equ: MoE class}) comprises expert networks and a gating mechanism. Their distinct roles are as follows:
% Recalling Equation~\eqref{equ: MoE class}, the depth-2 MoE comprises {\em two types} of subnetworks, expert networks and gating networks. Now we illustrate their different roles.
\begin{itemize}[leftmargin=2em]
    \item {\bf Expert networks} ($f^{(2,i)}$ in Layer 2): these components {\em efficiently approximate the $d$-dimensional local target function $f|_{U_i} \circ \phi_i^{-1}$ and the smooth chart map $\phi_i$}.
    They directly influence the overall approximation error and benefit from increased width $m$.
    
    \item {\bf Routing mechanism} (Layer 1 and gating in Layer 2): these components work together to {\em exactly assign each input to its correct expert $f^{(2,i)}$}. 
    The first MoE layer $h^{(1)}$ behaves like a dense model, approximating the smooth partition functions $\rho_i$'s. Then the Layer-2 gating network $g^{(2)}$ selects the expert corresponding to the region $U_i$ such that $\bx \in U_i$.
    This {\em exact} assignment is nontrivival, please refer to the proof for details. 
\end{itemize}

% The following corollary establish a stronger result for a more specific class of smooth compact manifolds, as illustrated in Example~\ref{example: manifold without boundary}.

\begin{corollary}[Special case, highly regular atlas]
\label{thm: 1-layer, special}
    Let $\cM$ be a compact, $d$-dimensional smooth manifold in $\bbR^D$, with a highly regular atlas $\{(U_i,\phi_i)\}_{i\in[E]}$ (Example~\ref{example: manifold without boundary}). Let the target function $f:\cM\to\bbR$.
    Then for any $m\geq \Omega\big(E^{2}\big)$, there exists a depth-$2$ MoE network  $\Psi\in\cH_{2,m}^{2,E}$, with $E$ experts per layer, each a $2$-layer $m$-width dense network,
    such that:
    \begin{align*}
        \|f-\Psi\|_{L_\infty(\cM)}
        \leq
        \max_{i\in[E]}\!\!\!\!\!\!\!\!\underbrace{\cE_{2,m}^{\rm FFN}\big(f|_{U_i}\circ\phi_i^{-1}\big)}_{\text{approximate a {\color{red!70!black} low-dimensional} function}}\!\!\!\!\!\!\!\!\leq\max_{i\in[E]}\ 
        \tilde{\cO}\Big(m^{-\frac{\color{red!70!black}\kappa(f|_{U_i})}{\color{red!70!black} d}\wedge\frac{1}{2}}\Big).
    \end{align*}
% Here, the term $\cE_{2,m}^{\rm FFN}\big(f|_{U_i}\circ\phi_i^{-1};[0,1]^r\big)$ admits the rate in Theorem \ref{thm: C^k approximation rate} with $D=r$ and $K=\kappa(f|_{U_i})$.
\end{corollary}

Compared to Theorem~~\ref{thm: 1-layer, general}, Corollary~\ref{thm: 1-layer, special} applies to a more structured setting in which the local coordinate maps $\phi_i$ are linear and thus thus do not require approximation. This eliminates the second term in the error bound, and reduces each expert to a 2-layer dense network.

{\bf Comparison with prior works}~\citet{shaham2018provable,chen2019efficient}.
These works show that {\em dense networks} can also efficiently approximate functions on low-dimensional manifolds.
However, their analyses require additional regularity assumptions on manifolds, which enables explicit constructions of coordinate charts and partition functions. 
In contrast, our Theorem~\ref{thm: 1-layer, general} does not require such explicit formulations, applying to a broader class of smooth regular manifolds. 
More importantly, MoEs offer a fundamental {\bf computational advantage}: while dense networks activate all parameters for every input, MoEs selectively activate only a single expert per input.
To achieve a comparable approximation accuracy, the number of activated parameters in dense networks is roughly $E$ times greater than that in MoEs, as dense networks must simultaneously approximate all $E$ subproblems.

\subsection{Practical Suggestions}

Beyond theoretical guarantees, our analysis also offers several practical suggestions into the design of MoE architectures.

{\bf Incorporating nonlinearity into gating is critical.}
Our theoretical results indicate that accurate input-to-expert assignment requires approximating the partition functions $\rho_i$, which are generally nonlinear.
Since standard gating function is linear and lacks the capacity to model nonlinear $\rho_i$, an additional MoE (or dense) layer is needed prior to gating to approximate $\rho_i$.
If the router incorporates sufficient nonlinearity, e.g., a two-layer ReLU routing network, it can directly model complex partitions, reducing the depth and number of parameters. For instance, in Theorem~\ref{thm: 1-layer, general}, the required depth will reduce from $2$ to $1$, because the nonlinearity in router eliminates the need for a preceding MoE layer. Similarly, in Theorem~\ref{thm: deep MoE, general}, the required depth will reduce from $2L$ to $L$. 
Therefore, a more direct and potentially more efficient alternative is to {\bf incorporate nonlinearity directly into the gating network}, eliminating the need for a preceding MoE layer.
This observation is consistent with recent empirical findings~\citep{zuo2021taming,liu2022gating,nguyen2024statistical,akbarian2024quadratic,le2024mixture}, which demonstrate that nonlinear gating functions improve MoE performance.

{\bf Alternating MoE architectures with equivalent expressive power.}
In our construction, the first MoE layer actually serves as a standard dense layer with width $\Omega(E^2)$ to approximate the partition functions $\rho_i$.
This insight motivates alternative architectures with comparable expressive power:
(i) MoE-dense alternating networks.
A natural variant consists of alternating dense and MoE layers, e.g., $h^{\rm moe} \circ h^{\rm dense}$. 
This design extends naturally to deeper architectures. 
This design has been adopted in practice by GShard~\citep{lepikhin2020gshard} and GLAM~\citep{du2022glam}.
(ii) Shared + routed experts per MoE layer.
Another common MoE variant incorporates one shared expert alongside $E$ routed experts. 
This structure is empirically adopted in modern MoE architectures such as Qwen2~\citep{yang2024qwen2} and DeepSeek~\citep{liu2024deepseek}.

{\bf Low-dimensional expert networks via autoencoding.}
Our analysis also suggests a more structured and interpretable design for expert networks in MoE.
Typically, each expert is implemented as a dense network with input dimension $D$ and width $\mathcal{O}(D)$, resulting in $\mathcal{O}(D^2)$ parameters.
However, our theory motivates replacing each expert $f^{(2l,i)}$ with a composition $f_{\mathrm{low}} \circ \mathrm{Enc}$, where:
$\mathrm{Enc}: \mathbb{R}^D \to \mathbb{R}^d$ is an encoder approximating the smooth coordinate chart $\phi: U_i \to [0,1]^d$,
$f_{\mathrm{low}}: \mathbb{R}^d \to \mathbb{R}$ is a low-dimensional dense network approximating $f_{l,i} \circ \phi^{-1}$.
This design reduces the number of trainable parameters in each expert to $\#(\mathrm{Enc}) + \mathcal{O}(d^2)$, which is significantly smaller than $\mathcal{O}(D^2)$ when $d \ll D$.
Moreover, this decomposition aligns with the manifold structure of the target function, improving interpretability.
To support encoder learning, one can incorporate a standard reconstruction loss $\mathbb{E}_{\bx} \| \mathrm{Dec}(\mathrm{Enc}(\bx)) - \bx \|_2^2$ using a decoder $\mathrm{Dec}$.
We leave empirical validation of this theoretically motivated architecture for future work.

\section{Theory for Multi-layer MoE Networks}\label{section: deep MoE}

{\bf Piecewise functions as multiple tasks.}
Modern LLMs are capable of performing a wide range of tasks, such as mathematics, logical reasoning, language understanding, and code generation.
From a mathematical perspective, each task can be viewed as a function defined on a task-specific input domain. 
In practice, these tasks often differ, which are supported on distinct regions $\Omega_1,\cdots,\Omega_N$, with distinct corresponding tasks $f_i:\Omega_i\to\bbR$. 
Therefore, performing $N$ tasks can be naturally modeled as approximating a piecewise function: $f(\bx)=f_i(\bx), \text{ if } \bx\in \Omega_i,i\in[N]$. 
While each $f_i$ may be high-order smooth within its region $\Omega_i$, the global function $f$ may exhibit only low-order smoothness at the interfaces between adjacent regions. 
% (For example, $f_i(x)=(x-2i)^2$ if $x\in\Omega_i=[2i-1,2i+1]$.)

{\bf Key question.}
Theorem~\ref{thm: 1-layer, general} shows that a depth-2 MoE network with $E$ experts per layer can efficiently approximate a piecewise function $f$ comprising $E$ pieces ($f|_{U_i},i\in[E]$) (each handled by a different expert).
A natural question then arises for deep MoE networks: 
\begin{center}\em
    How many distinct pieces can be efficiently modeled by a deep MoE network?
\end{center}

{\bf A naive limitation.}
As illustrated above, it is intuitive to associate each expert with a distinct task.
A depth-$L$ MoE with $E$ experts per layer contains $\cO(LE)$ experts in total, implying a capacity to model at most $\cO(LE)$ distinct regions if each expert is used independently.

{\bf Overview of our result: beyond the naive limitation.}
Surprisingly, this limitation can be overcome when the target function exhibits {\em compositional sparsity}.
We will show that:
\begin{center}
    \em Depth-$\mathcal{O}(L)$ MoE networks with $E$ experts per layer can efficiently approximate a piecewise function comprising $E^L$ pieces, provided the function satisfies a compositional sparsity structure.
\end{center}
This demonstrates that MoE networks can model an {\bf exponential number} of structured tasks (far surpassing the native limitation $\cO(LE)$) by exploiting structured sparsity in the function.

\subsection{Piecewise function with compositional sparsity}

\begin{wrapfigure}{r}{0.4\textwidth}
    \vspace{-1.5cm}
    \centering
    \includegraphics[width=0.3\textwidth]{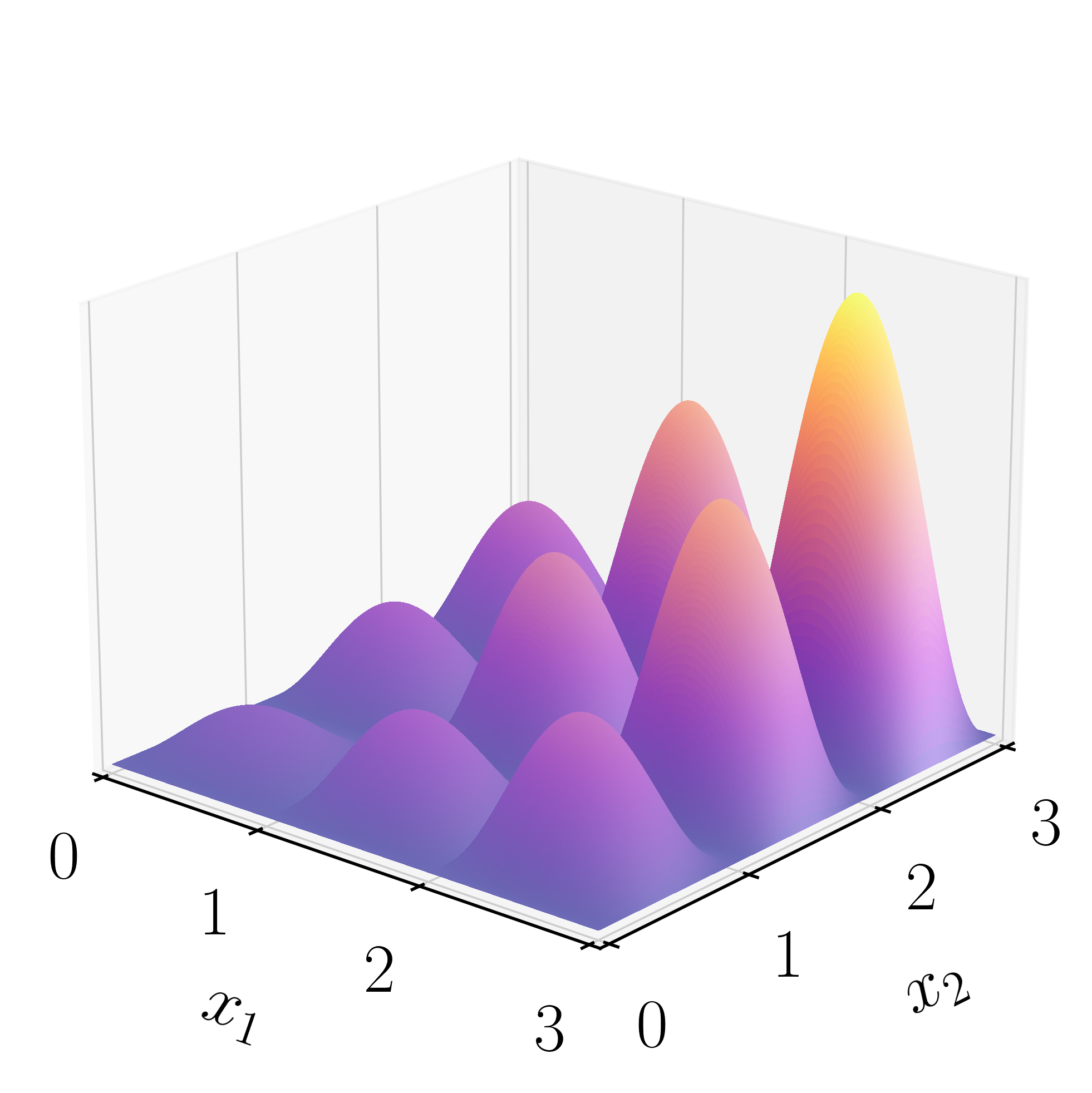}
    \caption{\small Illustration of Eq.~\eqref{example: piecewise function, 1-dim}: a piecewise function $f$  with compositional sparsity on $3^2 = 9$ unit cubes. The function $f(\bx) = f_{1,i_1}(x_1) f_{2,i_2}(x_2)$ is composed from 6 subfunctions: $f_{1,i}(z) = i(i - 1 - z)(z - i)$ and $f_{2,i}(z) = i(i - 1 - z)^2(z - i)^2$ on $z \in [i - 1, i]$ for $i \in [3]$.
     Although $f$ is smooth within each region, it is only $0$-order continuous on $[0,3] \times [0,3]$.}
     \vspace{-1.5cm}
     \label{fig: piecewise function}
\end{wrapfigure}
\paragraph{Warm-up: Piecewise function on $E^L$ unit cubes with compositional sparsity.}
Let the domain be $\cM=[0,E]^L$, naturally partitioned into $E^L$ unit cubes.
Consider the following target function:
\begin{equation}\label{example: piecewise function, 1-dim}
\begin{gathered}
    f(\bx)=(f_{1,i_1}(x_1),f_{2,i_2}(x_2),\cdots,f_{L,i_L}(x_L)),\\ 
\text{ where }
i_l\in\{j\in[E]:x_l\in[j-1,j]\},\ \forall l\in[L].
\end{gathered}
\end{equation}

Here, each subfunction (subtask) $f_{l,i}$ is defined on the interval $U_{l,i}=[i-1,i]$, and $f$ is defined piecewise over $E^L$ regions: $U_{i_1} \times \cdots \times U_{i_L}$, where $i_{l}\in[E],l\in[L]$).

The function $f$ in Eq.~\eqref{example: piecewise function, 1-dim} exhibits both:
\begin{itemize}[leftmargin=2em]
    \item {\bf Sparsity}: each subfucntion $f_{l,i}$ depends only on the coordinate $x_l$, not on the full input $\bx$.
    \item {\bf Compositionality}: the function $f$ is a hierarchical composition of $L$ selective subfunctions.
\end{itemize}
Notably, although there are only $LE$ subfunctions, their composition yields $E^L$ distinct functions across the domain.

% \begin{figure}
%     \centering
%     \includegraphics[width=0.3\textwidth]{figures/piecewise-function.png}
%     \caption{Example for Eq.~\eqref{fig: piecewise function}. A piecewise function on $2^3$ unit cubes with compositional sparsity. This function has $9$ smooth pieces, and only has du}
%     \label{fig:enter-label}
% \end{figure}

% Notably, in this case, the function $f$ may only has low-order continuity at the boundaries between the unit cubes, such as $\kappa(f)=0$. Therefore, approximating the entire function is hard. However, for each subfunction $f_{l,i}$, they may be high-order smoothness, with $\kappa(f_{l,i})\gg 1$.
% Therefore, the efficient method is to properly decouple this sparsity structure and use subnetworks approximate these simple subfunctions.

\paragraph{\bf Product manifold.}
We now extend the above formulation to on general manifolds.
Consider a product manifold $\cM=\cM_1\times\cM_2\times\cdots\times\cM_L$, where each $\cM_l$ is a compact $d_l$-dimensional manifold in $\bbR^D$, and $\sum_{l=1}^L d_l\leq D$.
Each input $\bx\in\cM$ can be written as $\bx=(\bx_1,\cdots,\bx_L)$ with $\bx_l\in\cM_l$.

By compactness, each submanifold $\cM_l$ admits a finite smooth atlas $\{(U_{l,i},\phi_{l,i})\}_{i\in[E_l]}$ and an associated partition of unity $\{\rho_{l_i}\}_{i\in[E_l]}$ (by Theorem~\ref{thm: partition of unity}).
Without loss of generality, we can let $E_1=\cdots=E_{L}$, denoted by $E$.

% \begin{figure}[!ht]
% % {r}{0.4\textwidth}
%     \vspace{-.4cm}
%     \centering
%     \includegraphics[width=0.3\textwidth]{figures/piecewise-function.png}
%     \caption{\small Illustration of Eq.~\eqref{example: piecewise function, 1-dim}: a piecewise function $f$  with compositional sparsity on $3^2 = 9$ unit cubes. The function $f(\bx) = f_{1,i_1}(x_1) f_{2,i_2}(x_2)$ is composed from 6 subfunctions: $f_{1,i}(z) = i(i - 1 - z)(z - i)$ and $f_{2,i}(z) = i(i - 1 - z)^2(z - i)^2$ on $z \in [i - 1, i]$ for $i \in [3]$.
%      Although $f$ is smooth within each region, it is only $0$-order continuous on $[0,3] \times [0,3]$.}
%      \vspace{-.2cm}
%     \label{fig: piecewise function}
%      % \vspace{-1.cm}
% \end{figure}

\paragraph{\bf General form: piecewise function on product manifold with compositional sparsity.}
Consider the target function class admit the form\footnote{To ensure $f$ is well-defined on overlapping charts, we assume $f_{l,i}(\bx_l)=f_{l,j}(\bx_l)$, $\forall\bx_l\in U_i\cap U_j,i\ne j$.}:
\begin{equation}\label{equ: piecewise function, general}
\begin{gathered}
    f(\bx)=f_{\rm out}(f_{1,i_1}(\bx_1),f_{2,i_2}(\bx_2),\cdots,f_{L,i_L}(\bx_L)),\quad \text{where}
    \\
    i_l\in\{j\in[E]:\bx_l\in U_{l,j}\},\ \forall l\in[L].
\end{gathered}
\end{equation}
Here, each subfunction (subtask) $f_{l,i}$ is defined on the local subregion $U_{l,i}\subset \cM_l$, and $f_{\rm out}$ composes their outputs. This extends Eq.~\eqref{example: piecewise function, 1-dim} from Euclidean coordinates $x_l\in[0,E]$ to manifold-based coordinates $\bx_l\in\cM_l$.
Since $f_{\rm out}$ can typically be approximated by a dense neural network, we assume $f_{\rm out}={\rm id}$ (the identity map) for simplicity.

We now illustrate this formulation with a concrete example.

\begin{table}[!ht]
\centering
\caption{\small Semantic interpretation of subregions in Example~\ref{example: piecewise function, practice}}
\begin{tabular}{c|c}
\hline\hline
 $\cM_1$: \texttt{Math domain} & $\cM_2$: \texttt{Language domain}   \\ \hline
 $U_{1,1}$: \texttt{Geometry} & $U_{2,1}$: \texttt{English} \\ 
 $U_{1,2}$: \texttt{Algebra}  & $U_{2,2}$: \texttt{French} \\
 $U_{1,3}$: \texttt{Analysis} & $U_{2,3}$: \texttt{German}
 \\\hline\hline
\end{tabular}
\label{table: multi tasks example}
\end{table}

\begin{example}\label{example: piecewise function, practice}
    Let $\cM=\cM_1\times\cM_2$, where each $\cM_l$ is partitioned into three subregions: $\cM_1=U_{1,1}\cup U_{1,2}\cup U_{1,3}$, $\cM_2=U_{2,1}\cup U_{2,2}\cup U_{2,3}$, with the interpretations given in Table~\ref{table: multi tasks example}.
    Each subfunction $f_{1,i}$ solves the a specific type of math problem (e.g., \texttt{geometry}), while each $f_{2,i}$ handles text comprehension in a specific language (e.g., \texttt{English}). 
    The full function $f$ defined via Eq.~\eqref{equ: piecewise function, general} encodes {\bf $3 \times 3 = 9$} compositional tasks of the form:
    \begin{center}
        ``Understand and solve the [\texttt{language type}] [\texttt{math type}] problem it''.
    \end{center}
    For example, if $\bx_1\in U_{1,1}$ and $\bx_2\in U_{2,1}$, then $f$ corresponds to the task ``understand and solve the \texttt{English geometry} problem''.
\end{example}

\subsection{Theoretical results and insights}

We now present our main theoretical result regarding the expressive power of deep MoE networks for approximating piecewise functions with compositional sparsity.

\begin{theorem}[Main result]\label{thm: deep MoE, general}
    Let the target function $f$ be of the form~\eqref{equ: piecewise function, general}, which comprises $E^L$ pieces.
    For each $l\in[L]$, assume that the atlas $\{(U_{l,i}, \phi_{l,i})\}_{i\in [E]}$ of $\cM_l$ is regular (Definition~\ref{def: regular manifold}).
    Then there exists a depth-$2L$ MoE network $\Psi\in\cH_{3,m}^{2L,E}$ with
    $m\geq\Omega\big(E^{2}\big)$,
    such that:
    \begin{align*}
        &\|f-\Psi\|_{L_\infty(\cM)}
        \leq\max_{l\in[L]}\max_{i\in[E]} \cE_{3,m}^{\rm FFN}\big(f_{l,i}\big)
        \\\leq&
        \max_{l\in[L]}\max_{i\in[E]}\underbrace{\cE_{2,m}^{\rm FFN}\big(f_{l,i}\circ\phi_{l,i}^{-1}\big)}_{\text{approximate a {\color{red!70!black} low-dimensional} function}}
        +\max_{l\in[L]}\max_{i\in[E_{l}]}  \!\!\!\underbrace{\cE_{2,m}^{\rm FFN}\big(\phi_{l,i}\big)}_{\text{approximate a {\color{red!70!black} smooth} map}}\!\!\!\!\norm{f_{l,i}\circ \phi_{l,i}^{-1}}_{\cC^1([0,1]^{d_l})}
        \\\leq&
         \max_{l\in[L]}\max_{i\in [E]}\tilde{\cO}\Big(m^{-\frac{\kappa(f_{l,i})}{d_l}\wedge\frac{1}{2}}\Big).
    \end{align*}
% Here, the term $\cE_{2,m}^{\rm FFN}\big(f|_{U_{l,i}}\circ\phi_{l,i}^{-1}\big)$ admits the rate in Theorem~\ref{thm: C^k approximation rate} with $D=r$ and $K=\kappa(f|_{U_{l,i}}\circ\phi_{l,i}^{-1})$; 
% the term $\cE_{2,m}^{\rm FFN}\big(\phi_{l,i}\big)$ admits the rate in Theorem \ref{thm: C^k approximation rate} with $D=d$ and $K\gg d$.
\end{theorem}

Theorem~\ref{thm: deep MoE, general} establishes that a depth-$2L$ MoE network with $E$ experts per layer can efficiently approximate a piecewise function with $E^L$ pieces, provided the function exhibits compositional sparsity.
The approximation error consists of two components: 
(i) approximation of local low-dimensional subfunctions $f_{l,i}\circ\phi_{l,i}^{-1}$;
(ii) approximation of smooth coordinate maps $\phi_{l,i}$.
The resulting approximation rate is $\max_{l\in[L]}\max_{i\in [E]}\tilde{\cO}\Big(m^{-\frac{\kappa(f_{l,i})}{d_l}\wedge\frac{1}{2}}\Big)$,
which avoids the curse of dimenisonality. In the special case $L=1$, this result recovers Theorem~\ref{thm: deep MoE, general}.

% The proof of Theorem~\ref{thm: deep MoE, general} is referred to Appendix~[sss]. Here, we draw the key insight from the proof as follows, which is highly intuitive.

{\bf Key insight.}
The proof (deferred to Appendix~\ref{appendix: proofs for deep MoE}) reveals the following intuitions:

\begin{center} \em
    Each pair of MoE layers implements $E$ subtasks,\\ 
    and a depth-$2L$ architecture enables hierarchical composition of $E^L$ tasks.
\end{center}

% Specifically: 
\begin{itemize}[leftmargin=2em]
    \item For each $l \in [L]$, the $(2l - 1)$-st and $2l$-th layers approximate the $E$ subfunctions ${f_{l,i}}$,(${i \in [E]}$) defined on the manifold charts $U_{l,i}\in\cM_i$ and assign $\bx_l$ to their correct experts, following the same mechanism as in Theorem~\ref{thm: 1-layer, general}.
    \item Stacking $L$ such MoE blocks composes these representations hierarchically, approximating the final function $(f_{1,i_1}(\bx_1),\cdots,f_{L,i_L}(\bx_L))$.
    % Similar to Theorem~\ref{thm: 1-layer, general}, for each $l\in[L]$, expert networks $f_{(l,i)}^{\rm moe}$ in Layer $2l$ efficiently approximates the simple functions localized on $U_i$; and the network in Layer $2l-1$ and the gating network in Layer $2l$ can exactly assign the input to its correct expert. 
    % Therefore, the $(2l-1)$-th and $l$-th layers together achieve the efficient approximate of the $E$ subtasks $f_{l,i_l}(\bx_l)$, where $i_l\in\{j\in[E:\bx_l\in U_{l,i}\}$.

    % \item Based on the above efficient approximation of $E$ subtasks in each two layers, the deep composition structure of deep networks naturally implement the composition of these subtasks, forming the final piecewise function $(f_{1,i_1}(\bx_1),\cdots,f_{1,i_L}(\bx_L))$ in Equation~\eqref{equ: piecewise function, general}, which containing $E^L$ tasks.
\end{itemize}

{\bf Illustration via Example~\ref{example: piecewise function, practice}.}
In Example~\ref{example: piecewise function, practice} with $L=2$ and $E=3$ (3 subregions for each of math and language). 
Theorem~\ref{thm: deep MoE, general} shows that a depth-4 MoE network with 3 experts per layer can express all $3 \times 3 = 9$ tasks of the form ``understand and solve the [\texttt{language type}] [\texttt{math type}] problem''. Concretely,
\begin{itemize}[leftmargin=2em]
    \item Experts in Layer 2 (math) implement $f_{2,i}$ for $i=1,2,3$, corresponding to ``solving \texttt{geometry}/ \texttt{algebra}/\texttt{analysis} problems'', respectively..
    \item Experts in Layer 4 (language) implement $f_{4,i}$ for $i=1,2,3$, corresponding to ``understanding \texttt{English}/\texttt{French}/\texttt{German}'', respectively.
    \item Layer 1 and the gating in Layer 2 together perform routing for $\bx_1$; Layer 3 and the gating in Layer 4 together perform routing for $\bx_2$.
\end{itemize}

% $f_{(2,1)}^{\rm moe}$, $f_{(2,2)}^{\rm moe}$, $f_{(2,3)}^{\rm moe}$ approximate the subtasks ``solving the \texttt{geometry}, \texttt{algebra}, \texttt{analysis} problems'', respectively. 
% The first two layers together can assign $\bx_1$ to its correct expert $f_{(2,i)}^{\rm moe}$ such that $\bx_1\in U_{1,i}$'s.
% $f_{(4,1)}^{\rm moe}$, $f_{(4,2)}^{\rm moe}$, $f_{(4,3)}^{\rm moe}$ approximate the subtasks ``understaning the \texttt{English}, \texttt{French}, \texttt{German} text'', respectively. 
% The first two layers together can assign $\bx_2$ to its correct expert $f_{(4,i)}^{\rm moe}$ such that $\bx_2\in U_{2,i}$'s.
% Totally, the network achieves $9$ tasks ``understand the [\texttt{language type}] [\texttt{math type}] problem and solve it''.

% Layers 1 and 3 implement routing for $\bx_1$ and $\bx_2$, respectively.

\begin{theorem}[Warmup case]
\label{thm: deep MoE, warmup}
Let the target function $f$ be of the form~\eqref{example: piecewise function, 1-dim}.
Then there exists a depth-$2L$ MoE network $\Psi\in\cH_{2,m}^{2L,E}$ with $m\geq\Omega(E)$, such that:
\begin{align*}
    \|f-\Psi\|_{L_\infty(\cM)}\leq
    \max_{l\in[L]}\max_{i\in[E]}\!\!\!\!\!\! \underbrace{\cE_{2,m}^{\rm FFN}\big(f_{l,i}\big)}_{\text{approximate a {\color{red!70!black} 1-dimensional} function}}
    \!\!\!\!\!\!\leq\max_{l\in[L]}\max_{i\in[E]}\cO\Big(m^{-\kappa(f_{l,i})\wedge\frac{1}{2}}\Big).
\end{align*}
\end{theorem}

Compared to Theorem~\ref{thm: deep MoE, general}, this result requires only: (i) shallower expert networks (2-layer instead of 3-layer), (ii) smaller expert width ($m\geq \Omega(E)$ instead of $\Omega(E^2)$) due to the simple geometry (Euclidean cubes).

Although each subfunction may have high-order smoothness ($\kappa(f_{l,i}) \gg 1$), the composite function $f$ can exhibit only low-order smoothness (e.g., $\kappa(f)=0,1$) due to low-order regularity at the interfaces between adjacent regions.
As a result, directly approximating the global function $f$ is inefficient. 
A natural and efficient approach is to decompose the approximation problem into localized subproblems, each defined on a simple subregion with high regularity. Our constructive MoE networks can provably achieve this approach.

% In fact, It is straightforward to obtain a similar fine-grained analysis by adding weaker regularity conditions on the manifold, such as the reach condition~\citep{aamari2019estimating}. More examples are leaved for future works.
% Further refinements under weaker regularity assumptions on the manifold (e.g., positive reach~\citep{aamari2019estimating}) are left for future work.
% Notably, in this case, the function may only has low-order continuity at the boundaries between the unit cubes. Therefore, the function on the entire $\cM$ has a low smoothness, is hard to be approximated. However, by the , each network only need to approximate

\section{Experimental Validation}
\label{section: experiments}

To support our main theoretical results, we conduct two new experiments, each aligned with one of our key insights. The experimental details are shown in Appendix~\ref{appendix: experiments}.

{\bf Experiment I. Shallow MoEs for low-dimensional functions.} To validate our theoretical insight in Section~\ref{section: shallow MoE} (Theorem~\ref{thm: 1-layer, general}): shallow MoE networks can efficiently approximate functions supported on low-dimensional manifolds and {\em overcome the curse of dimensionality}.

Specifically, we consider the low-dimensional manifold $\mathcal{M}=\{\boldsymbol{x}\in\mathbb{R}^D:x_1^2+x_2^2=1;x_i=0,\forall i>2\}$ embedded in $\mathbb{R}^D$ with $D>2$. The target function is $f(\boldsymbol{x})=\sin(5x_1)+\cos(3x_2)$, defined on $\mathcal{M}$. As a model, we consider ``1-4-MoE'', a 1-layer MoE comprising 1 router and 4 experts, where each expert is a two-layer ReLU network with hidden width $10$. To validate whether MoE can overcome the curse of dimensionality, we vary the input dimension $D\in\{16,32,64,128\}$.

As shown in Table~\ref{tab: experiment I}, one can see that: as $D$ increases, the test error of MoE does not increase significantly and remains stable. This supports our insight that shallow MoEs efficiently approximate functions on low-dimensional manifolds and avoid the curse of dimensionality.

\vspace{-.2cm}

\begin{table}[!ht]
    \centering
    \caption{\small (Results of Experiment I) The test error of 1-4-MoE under different input dimensions $D$.}
    \begin{tabular}{c|c|c|c|c}
    \hline\hline
        input dim $D$ & $16$ & $32$ & $64$ & $128$ \\\hline
        test error & $3.40$\texttt{e-4} & $3.38$\texttt{e-4} & $3.17$\texttt{e-4} & $3.42$\texttt{e-4} \\\hline\hline
    \end{tabular}
    \label{tab: experiment I}
\end{table}

\vspace{-.1cm}

{\bf Experiment II. Deep MoEs for piecewise functions.}
To verify our theoretical insight in Section~\ref{section: deep MoE}: depth-$\cO(L)$ MoE networks with $E$ experts per layers can efficiently approximate piecewise functions with $E^L$ distinct pieces.

As defined in our Figure~\ref{fig: piecewise function}, we consider the piecewise function $f$ with compositional sparsity defined over $3^2 = 9$ unit cubes.
As the model, we consider ``2-3-MoE'' (a 2-layer MoE comprising 2 routing layers and 2 expert layers with 3 experts each); To illustrate the role of depth, we also consider a shallow ``1-6-MoE'', with comparable parameter count. Each expert is a two-layer ReLU FFN with hidden width $m\in\{16,32,64,128\}$. 
To validate whether 2-3-MoE and 1-6-MoE can approximate this target, we vary the hidden width $m$.

The results, shown in Table~\ref{tab: experiment II}, illustrate that: {\em (i)} As $m$ increases, 2-3-MoE achieves rapidly decreasing error. This supports that the depth-$2$ MoE with $3$ experts per layers can efficiently approximate this piecewise function with $3^2$ distinct pieces; {\em (ii)} In contrast, 1-6-MoE exhibits a performance plateau, revealing its limited expressive power. This highlights the crucial role of depth in modeling such compositional structures.

\vspace{-.2cm}

\begin{table}[!ht]
    \centering
    \caption{\small (Results of Experiment II) The test error of the 2-3-MoE and 1-6-MoE under different hidden width $m$.}
    \begin{tabular}{c|c|c|c|c}
    \hline\hline
    hidden width $m$ of experts & $16$ & $32$ & $64$ & $128$ \\\hline
    test error of 2-3-MoE & $8.32$\texttt{e-5}  & $1.41$\texttt{e-5} & $4.73$\texttt{e-6} & $2.59$\texttt{e-6} \\\hline
    test error of 1-6-MoE & $7.96$\texttt{e-5} & $2.17$\texttt{e-5} & $2.65$\texttt{e-5} & $4.60$\texttt{e-5} \\\hline\hline
    \end{tabular}
    \label{tab: experiment II}
\end{table}

\vspace{-.1cm}

\section{Conclusion and Future Work}
\label{section: conclusion}

% \vspace{-.1cm}

In this work, we provide a theoretical study of the expressive power of MoE networks for modeling structured complex tasks.
For shallow MoE networks, we show that they can efficiently approximate functions supported on low-dimensional manifolds, overcoming the curse of dimensionality.
For deep MoE networks, we establish that, when the target exhibits a compositional sparsity structure, they can approximate piecewise functions consisting of exponentially many distinct pieces, effectively modeling an exponential number of tasks.
% Future directions include:

{\bf Beyond compositional sparsity.}
Our analysis focuses on compositional sparsity, a structure that is both theoretically rich and practically relevant.
However, real-world problems may involve other forms of sparsity, such as group sparsity, graph sparsity, or temporal sparsity.
Characterizing the expressive power of deep MoE networks under these alternative structures is an important direction for future work.

{\bf Training dynamics analysis.}
While our study addresses the approximation capabilities of MoE networks, a critical open question is whether such expressive solutions can be found via training algorithms such as stochastic gradient descent.
Analyzing the training dynamics of MoE networks is considerably more challenging than in dense architectures, due to auxiliary load-balancing objectives and the use of non-differentiable Top-$K$ routing.

% \vspace{-.1cm}

\section*{Acknowledgments}

% \vspace{-.1cm}

Mingze Wang is supported by Young
Scientists (PhD) Fund of the National Natural Science Foundation of China (No. 124B2028).

% \bibliography{ref}
% \bibliographystyle{plainnat}

\newpage

\section*{NeurIPS Paper Checklist}

\begin{enumerate}

\item {\bf Claims}
    \item[] Question: Do the main claims made in the abstract and introduction accurately reflect the paper's contributions and scope?
    \item[] Answer: \answerYes{} % Replace by \answerYes{}, \answerNo{}, or \answerNA{}.
    \item[] Justification: We believe that the abstract and introduction reflect the contributions and scope of the paper.
    \item[] Guidelines:
    \begin{itemize}
        \item The answer NA means that the abstract and introduction do not include the claims made in the paper.
        \item The abstract and/or introduction should clearly state the claims made, including the contributions made in the paper and important assumptions and limitations. A No or NA answer to this question will not be perceived well by the reviewers. 
        \item The claims made should match theoretical and experimental results, and reflect how much the results can be expected to generalize to other settings. 
        \item It is fine to include aspirational goals as motivation as long as it is clear that these goals are not attained by the paper. 
    \end{itemize}

\item {\bf Limitations}
    \item[] Question: Does the paper discuss the limitations of the work performed by the authors?
    \item[] Answer: \answerYes{} % Replace by \answerYes{}, \answerNo{}, or \answerNA{}.
    \item[] Justification: In Section~\ref{section: conclusion}.
    \item[] Guidelines:
    \begin{itemize}
        \item The answer NA means that the paper has no limitation while the answer No means that the paper has limitations, but those are not discussed in the paper. 
        \item The authors are encouraged to create a separate "Limitations" section in their paper.
        \item The paper should point out any strong assumptions and how robust the results are to violations of these assumptions (e.g., independence assumptions, noiseless settings, model well-specification, asymptotic approximations only holding locally). The authors should reflect on how these assumptions might be violated in practice and what the implications would be.
        \item The authors should reflect on the scope of the claims made, e.g., if the approach was only tested on a few datasets or with a few runs. In general, empirical results often depend on implicit assumptions, which should be articulated.
        \item The authors should reflect on the factors that influence the performance of the approach. For example, a facial recognition algorithm may perform poorly when image resolution is low or images are taken in low lighting. Or a speech-to-text system might not be used reliably to provide closed captions for online lectures because it fails to handle technical jargon.
        \item The authors should discuss the computational efficiency of the proposed algorithms and how they scale with dataset size.
        \item If applicable, the authors should discuss possible limitations of their approach to address problems of privacy and fairness.
        \item While the authors might fear that complete honesty about limitations might be used by reviewers as grounds for rejection, a worse outcome might be that reviewers discover limitations that aren't acknowledged in the paper. The authors should use their best judgment and recognize that individual actions in favor of transparency play an important role in developing norms that preserve the integrity of the community. Reviewers will be specifically instructed to not penalize honesty concerning limitations.
    \end{itemize}

\item {\bf Theory assumptions and proofs}
    \item[] Question: For each theoretical result, does the paper provide the full set of assumptions and a complete (and correct) proof?
    \item[] Answer: \answerYes{} % Replace by \answerYes{}, \answerNo{}, or \answerNA{}.
    \item[] Justification: In Section~\ref{section: shallow MoE} and~\ref{section: deep MoE}; Appendix~\ref{appendix: proofs for shallow MoE} and~\ref{appendix: proofs for deep MoE}.
    \item[] Guidelines:
    \begin{itemize}
        \item The answer NA means that the paper does not include theoretical results. 
        \item All the theorems, formulas, and proofs in the paper should be numbered and cross-referenced.
        \item All assumptions should be clearly stated or referenced in the statement of any theorems.
        \item The proofs can either appear in the main paper or the supplemental material, but if they appear in the supplemental material, the authors are encouraged to provide a short proof sketch to provide intuition. 
        \item Inversely, any informal proof provided in the core of the paper should be complemented by formal proofs provided in appendix or supplemental material.
        \item Theorems and Lemmas that the proof relies upon should be properly referenced. 
    \end{itemize}

    \item {\bf Experimental result reproducibility}
    \item[] Question: Does the paper fully disclose all the information needed to reproduce the main experimental results of the paper to the extent that it affects the main claims and/or conclusions of the paper (regardless of whether the code and data are provided or not)?
    \item[] Answer: \answerNA{} % Replace by \answerYes{}, \answerNo{}, or \answerNA{}.
    \item[] Justification: This work is purely theoretical and does not involve experiments.
    \item[] Guidelines:
    \begin{itemize}
        \item The answer NA means that the paper does not include experiments.
        \item If the paper includes experiments, a No answer to this question will not be perceived well by the reviewers: Making the paper reproducible is important, regardless of whether the code and data are provided or not.
        \item If the contribution is a dataset and/or model, the authors should describe the steps taken to make their results reproducible or verifiable. 
        \item Depending on the contribution, reproducibility can be accomplished in various ways. For example, if the contribution is a novel architecture, describing the architecture fully might suffice, or if the contribution is a specific model and empirical evaluation, it may be necessary to either make it possible for others to replicate the model with the same dataset, or provide access to the model. In general. releasing code and data is often one good way to accomplish this, but reproducibility can also be provided via detailed instructions for how to replicate the results, access to a hosted model (e.g., in the case of a large language model), releasing of a model checkpoint, or other means that are appropriate to the research performed.
        \item While NeurIPS does not require releasing code, the conference does require all submissions to provide some reasonable avenue for reproducibility, which may depend on the nature of the contribution. For example
        \begin{enumerate}
            \item If the contribution is primarily a new algorithm, the paper should make it clear how to reproduce that algorithm.
            \item If the contribution is primarily a new model architecture, the paper should describe the architecture clearly and fully.
            \item If the contribution is a new model (e.g., a large language model), then there should either be a way to access this model for reproducing the results or a way to reproduce the model (e.g., with an open-source dataset or instructions for how to construct the dataset).
            \item We recognize that reproducibility may be tricky in some cases, in which case authors are welcome to describe the particular way they provide for reproducibility. In the case of closed-source models, it may be that access to the model is limited in some way (e.g., to registered users), but it should be possible for other researchers to have some path to reproducing or verifying the results.
        \end{enumerate}
    \end{itemize}

\item {\bf Open access to data and code}
    \item[] Question: Does the paper provide open access to the data and code, with sufficient instructions to faithfully reproduce the main experimental results, as described in supplemental material?
    \item[] Answer: \answerNo{} % Replace by \answerYes{}, \answerNo{}, or \answerNA{}.
    \item[] Justification: The code or data of the experiments are simple and easy to reproduce following the description in the paper.
    \item[] Guidelines:
    \begin{itemize}
        \item The answer NA means that paper does not include experiments requiring code.
        \item Please see the NeurIPS code and data submission guidelines (\url{https://nips.cc/public/guides/CodeSubmissionPolicy}) for more details.
        \item While we encourage the release of code and data, we understand that this might not be possible, so “No” is an acceptable answer. Papers cannot be rejected simply for not including code, unless this is central to the contribution (e.g., for a new open-source benchmark).
        \item The instructions should contain the exact command and environment needed to run to reproduce the results. See the NeurIPS code and data submission guidelines (\url{https://nips.cc/public/guides/CodeSubmissionPolicy}) for more details.
        \item The authors should provide instructions on data access and preparation, including how to access the raw data, preprocessed data, intermediate data, and generated data, etc.
        \item The authors should provide scripts to reproduce all experimental results for the new proposed method and baselines. If only a subset of experiments are reproducible, they should state which ones are omitted from the script and why.
        \item At submission time, to preserve anonymity, the authors should release anonymized versions (if applicable).
        \item Providing as much information as possible in supplemental material (appended to the paper) is recommended, but including URLs to data and code is permitted.
    \end{itemize}

\item {\bf Experimental setting/details}
    \item[] Question: Does the paper specify all the training and test details (e.g., data splits, hyperparameters, how they were chosen, type of optimizer, etc.) necessary to understand the results?
    \item[] Answer: \answerYes{} % Replace by \answerYes{}, \answerNo{}, or \answerNA{}.
    \item[] Justification: In Section~\ref{section: experiments} and Appendix~\ref{appendix: experiments}.
    \item[] Guidelines:
    \begin{itemize}
        \item The answer NA means that the paper does not include experiments.
        \item The experimental setting should be presented in the core of the paper to a level of detail that is necessary to appreciate the results and make sense of them.
        \item The full details can be provided either with the code, in appendix, or as supplemental material.
    \end{itemize}

\item {\bf Experiment statistical significance}
    \item[] Question: Does the paper report error bars suitably and correctly defined or other appropriate information about the statistical significance of the experiments?
    \item[] Answer: \answerNo{} % Replace by \answerYes{}, \answerNo{}, or \answerNA{}.
    \item[] Justification: The approximation error is deterministic and there is no need to consider the
error bars here..
    \item[] Guidelines:
    \begin{itemize}
        \item The answer NA means that the paper does not include experiments.
        \item The authors should answer "Yes" if the results are accompanied by error bars, confidence intervals, or statistical significance tests, at least for the experiments that support the main claims of the paper.
        \item The factors of variability that the error bars are capturing should be clearly stated (for example, train/test split, initialization, random drawing of some parameter, or overall run with given experimental conditions).
        \item The method for calculating the error bars should be explained (closed form formula, call to a library function, bootstrap, etc.)
        \item The assumptions made should be given (e.g., Normally distributed errors).
        \item It should be clear whether the error bar is the standard deviation or the standard error of the mean.
        \item It is OK to report 1-sigma error bars, but one should state it. The authors should preferably report a 2-sigma error bar than state that they have a 96\% CI, if the hypothesis of Normality of errors is not verified.
        \item For asymmetric distributions, the authors should be careful not to show in tables or figures symmetric error bars that would yield results that are out of range (e.g. negative error rates).
        \item If error bars are reported in tables or plots, The authors should explain in the text how they were calculated and reference the corresponding figures or tables in the text.
    \end{itemize}

\item {\bf Experiments compute resources}
    \item[] Question: For each experiment, does the paper provide sufficient information on the computer resources (type of compute workers, memory, time of execution) needed to reproduce the experiments?
    \item[] Answer: \answerYes{} % Replace by \answerYes{}, \answerNo{}, or \answerNA{}.
    \item[] Justification: In Appendix~\ref{appendix: experiments}.
    \item[] Guidelines:
    \begin{itemize}
        \item The answer NA means that the paper does not include experiments.
        \item The paper should indicate the type of compute workers CPU or GPU, internal cluster, or cloud provider, including relevant memory and storage.
        \item The paper should provide the amount of compute required for each of the individual experimental runs as well as estimate the total compute. 
        \item The paper should disclose whether the full research project required more compute than the experiments reported in the paper (e.g., preliminary or failed experiments that didn't make it into the paper). 
    \end{itemize}
    
\item {\bf Code of ethics}
    \item[] Question: Does the research conducted in the paper conform, in every respect, with the NeurIPS Code of Ethics \url{https://neurips.cc/public/EthicsGuidelines}?
    \item[] Answer: \answerYes{} % Replace by \answerYes{}, \answerNo{}, or \answerNA{}.
    \item[] Justification: We have confirmed that the research is conducted with the NeurIPS Code of Ethics.
    \item[] Guidelines:
    \begin{itemize}
        \item The answer NA means that the authors have not reviewed the NeurIPS Code of Ethics.
        \item If the authors answer No, they should explain the special circumstances that require a deviation from the Code of Ethics.
        \item The authors should make sure to preserve anonymity (e.g., if there is a special consideration due to laws or regulations in their jurisdiction).
    \end{itemize}

\item {\bf Broader impacts}
    \item[] Question: Does the paper discuss both potential positive societal impacts and negative societal impacts of the work performed?
    \item[] Answer: \answerNA{} % Replace by \answerYes{}, \answerNo{}, or \answerNA{}.
    \item[] Justification: \answerNA{}
    \item[] Guidelines:
    \begin{itemize}
        \item The answer NA means that there is no societal impact of the work performed.
        \item If the authors answer NA or No, they should explain why their work has no societal impact or why the paper does not address societal impact.
        \item Examples of negative societal impacts include potential malicious or unintended uses (e.g., disinformation, generating fake profiles, surveillance), fairness considerations (e.g., deployment of technologies that could make decisions that unfairly impact specific groups), privacy considerations, and security considerations.
        \item The conference expects that many papers will be foundational research and not tied to particular applications, let alone deployments. However, if there is a direct path to any negative applications, the authors should point it out. For example, it is legitimate to point out that an improvement in the quality of generative models could be used to generate deepfakes for disinformation. On the other hand, it is not needed to point out that a generic algorithm for optimizing neural networks could enable people to train models that generate Deepfakes faster.
        \item The authors should consider possible harms that could arise when the technology is being used as intended and functioning correctly, harms that could arise when the technology is being used as intended but gives incorrect results, and harms following from (intentional or unintentional) misuse of the technology.
        \item If there are negative societal impacts, the authors could also discuss possible mitigation strategies (e.g., gated release of models, providing defenses in addition to attacks, mechanisms for monitoring misuse, mechanisms to monitor how a system learns from feedback over time, improving the efficiency and accessibility of ML).
    \end{itemize}
    
\item {\bf Safeguards}
    \item[] Question: Does the paper describe safeguards that have been put in place for responsible release of data or models that have a high risk for misuse (e.g., pretrained language models, image generators, or scraped datasets)?
    \item[] Answer: \answerNA{} % Replace by \answerYes{}, \answerNo{}, or \answerNA{}.
    \item[] Justification: \answerNA{}
    \item[] Guidelines:
    \begin{itemize}
        \item The answer NA means that the paper poses no such risks.
        \item Released models that have a high risk for misuse or dual-use should be released with necessary safeguards to allow for controlled use of the model, for example by requiring that users adhere to usage guidelines or restrictions to access the model or implementing safety filters. 
        \item Datasets that have been scraped from the Internet could pose safety risks. The authors should describe how they avoided releasing unsafe images.
        \item We recognize that providing effective safeguards is challenging, and many papers do not require this, but we encourage authors to take this into account and make a best faith effort.
    \end{itemize}

\item {\bf Licenses for existing assets}
    \item[] Question: Are the creators or original owners of assets (e.g., code, data, models), used in the paper, properly credited and are the license and terms of use explicitly mentioned and properly respected?
    \item[] Answer: \answerNA{} % Replace by \answerYes{}, \answerNo{}, or \answerNA{}.
    \item[] Justification: \answerNA{}
    \item[] Guidelines:
    \begin{itemize}
        \item The answer NA means that the paper does not use existing assets.
        \item The authors should cite the original paper that produced the code package or dataset.
        \item The authors should state which version of the asset is used and, if possible, include a URL.
        \item The name of the license (e.g., CC-BY 4.0) should be included for each asset.
        \item For scraped data from a particular source (e.g., website), the copyright and terms of service of that source should be provided.
        \item If assets are released, the license, copyright information, and terms of use in the package should be provided. For popular datasets, \url{paperswithcode.com/datasets} has curated licenses for some datasets. Their licensing guide can help determine the license of a dataset.
        \item For existing datasets that are re-packaged, both the original license and the license of the derived asset (if it has changed) should be provided.
        \item If this information is not available online, the authors are encouraged to reach out to the asset's creators.
    \end{itemize}

\item {\bf New assets}
    \item[] Question: Are new assets introduced in the paper well documented and is the documentation provided alongside the assets?
    \item[] Answer: \answerNA{} % Replace by \answerYes{}, \answerNo{}, or \answerNA{}.
    \item[] Justification: \answerNA{}
    \item[] Guidelines:
    \begin{itemize}
        \item The answer NA means that the paper does not release new assets.
        \item Researchers should communicate the details of the dataset/code/model as part of their submissions via structured templates. This includes details about training, license, limitations, etc. 
        \item The paper should discuss whether and how consent was obtained from people whose asset is used.
        \item At submission time, remember to anonymize your assets (if applicable). You can either create an anonymized URL or include an anonymized zip file.
    \end{itemize}

\item {\bf Crowdsourcing and research with human subjects}
    \item[] Question: For crowdsourcing experiments and research with human subjects, does the paper include the full text of instructions given to participants and screenshots, if applicable, as well as details about compensation (if any)? 
    \item[] Answer: \answerNA{} % Replace by \answerYes{}, \answerNo{}, or \answerNA{}.
    \item[] Justification: \answerNA{}
    \item[] Guidelines:
    \begin{itemize}
        \item The answer NA means that the paper does not involve crowdsourcing nor research with human subjects.
        \item Including this information in the supplemental material is fine, but if the main contribution of the paper involves human subjects, then as much detail as possible should be included in the main paper. 
        \item According to the NeurIPS Code of Ethics, workers involved in data collection, curation, or other labor should be paid at least the minimum wage in the country of the data collector. 
    \end{itemize}

\item {\bf Institutional review board (IRB) approvals or equivalent for research with human subjects}
    \item[] Question: Does the paper describe potential risks incurred by study participants, whether such risks were disclosed to the subjects, and whether Institutional Review Board (IRB) approvals (or an equivalent approval/review based on the requirements of your country or institution) were obtained?
    \item[] Answer: \answerNA{} % Replace by \answerYes{}, \answerNo{}, or \answerNA{}.
    \item[] Justification: \answerNA{}
    \item[] Guidelines:
    \begin{itemize}
        \item The answer NA means that the paper does not involve crowdsourcing nor research with human subjects.
        \item Depending on the country in which research is conducted, IRB approval (or equivalent) may be required for any human subjects research. If you obtained IRB approval, you should clearly state this in the paper. 
        \item We recognize that the procedures for this may vary significantly between institutions and locations, and we expect authors to adhere to the NeurIPS Code of Ethics and the guidelines for their institution. 
        \item For initial submissions, do not include any information that would break anonymity (if applicable), such as the institution conducting the review.
    \end{itemize}

\item {\bf Declaration of LLM usage}
    \item[] Question: Does the paper describe the usage of LLMs if it is an important, original, or non-standard component of the core methods in this research? Note that if the LLM is used only for writing, editing, or formatting purposes and does not impact the core methodology, scientific rigorousness, or originality of the research, declaration is not required.
    %this research? 
    \item[] Answer: \answerNA{} % Replace by \answerYes{}, \answerNo{}, or \answerNA{}.
    \item[] Justification: This paper does not use LLMs as an important, original, or non-standard component of its core methods.
    LLM is used only for writing, editing, or formatting purposes.
    \item[] Guidelines:
    \begin{itemize}
        \item The answer NA means that the core method development in this research does not involve LLMs as any important, original, or non-standard components.
        \item Please refer to our LLM policy (\url{https://neurips.cc/Conferences/2025/LLM}) for what should or should not be described.
    \end{itemize}

\end{enumerate}

\newpage
\appendix

\begin{center}
    \noindent\rule{\textwidth}{4pt} \vspace{-0.2cm}
    % \noindent\rule{\textwidth}{1.2pt} \vspace{-0.25cm}
    \LARGE \textbf{Appendix} % \\ ~\\[-0.5cm]
    \noindent\rule{\textwidth}{1.2pt}
\end{center}

\startcontents[sections]
\printcontents[sections]{l}{1}{\setcounter{tocdepth}{2}}

\vspace{1.cm}

\section{Proofs in Section~\ref{section: shallow MoE}}
\label{appendix: proofs for shallow MoE}

\subsection{Further Introduction to Manifolds}

Compact smooth manifolds $\cM\subset\bbR^D$ are known to have strong regularity properties.

We begin by introducing the definition of the reach, a geometric quantity affected by two factors: the curvature of the manifold and the width of the narrowest bottleneck-like structure of $\cM$.

\begin{definition}[reach]
Let $\cA(\cM):=\{\bz\in\bbR^D:\exists \bp\ne\bq\in\cM,\|\bp-\bz\|_2=\|\bq-\bz\|_2=\inf\limits_{\by\in\cM}\|\by-\bz\|_2\}$ be the set of points that have at least two nearest neighbors on $\cM$. 
Then the reach of $\cM$ is defined as $\tau_\cM:=\inf\limits_{\bz\in\cM,\by\in\cA(\cM)}\|\by-\bz\|_2$.
\end{definition}

The following lemma has established that a compact smooth manifold has positive reach.

\begin{lemma}[\citet{federer1959curvature}]
Let $\cM$ be a compact smooth manifold in $\bbR^D$. Then it has positive reach, i.e., $\tau_\cM>0$.
\end{lemma}

If $\cM$ has the reach greater than $\tau>0$, then intuitively, one can roll freely a ball of radius $\tau$ around it. This ensures the existence of a well-structured atlas of compact smooth manifold in Example~\ref{example: manifold without boundary}:

{\bf Highly regular atlas of compact smooth manifold $\cM$.}~\citep{chen2019efficient}
Let $r>0$ and consider the open cover $\{\bbB(\bx;r)\}_{\bx\in\cM}$ of $\cM$, where $\bbB(\bx; r)$ denotes the $\ell_2$ Euclidean ball of radius $r$ centered at $\bx$.
Since $\cM$ is compact, there exists a finite subcover such that $\cM\subset\cup_{i\in[E]}\bbB(\bc_i;r)$, where $\bc_i\in\cM$.
Now we pick the radius $r<\tau_\cM/2$ such that 
\begin{align*}
    U_i=\cM\cap\bbB(\bc_i;r)
\end{align*} 
is diffeomorphic to a ball in $\bbR^d$~\citep{niyogi2008finding}.
We denote the tangent space at $\bc_i$ as $\cT_{\bc_i}(\cM)={\rm span}(\bv_{i,1},\cdots,\bv_{i,d})$, where  $\{\bv_{i,1},\cdots,\bv_{i,d}\}$ are orthonormal basis. 
Define the matrix $\bV_i=[\bv_{i,1},\cdots,\bv_{i,d}]\in\bbR^{D\times d}$ and construct the chart map as:
\begin{align*}
    \phi_i(\bx)=b_i(\bV_i^\top(\bx-\bc_i)+\bs_i)\in[0,1]^d,\quad\forall \bx\in U_i,
\end{align*}
where $b_i\in(0,1]$ is a scaling factor and $s_i$ is a translation vector chosen such that  $\phi_i(\bx)\in[0,1]^d$.
Then $\{(U_i,\phi_i)\}_{i\in[E]}$ is a smooth atlas on $\cM$. We call it highly regular, since (i) each $U_i$ is diffeomorphic to a ball in $\bbR^d$; (ii) each chart map $\phi_i$ is a linear function, implying infinite smoothness.

\subsection{Proof of Theorem~\ref{thm: 1-layer, general}}
\label{subappendix: proof shallow, main}

For simplicity, we use the notation $\norm{\cdot}$ to denote the $L_{\infty}$ norm in the proof.

\underline{\bf Routing mechanism (Layer 1 and gating in Layer 2).}

{\em Proof sketch:} These components work together to {\em exactly assign each input $\bx$ to its correct expert $f^{(2,i)}$}. Specifically, the first MoE layer $h^{(1)}$ behaves like a dense model, approximating the smooth partition functions $\rho_i$'s. Then the Layer-2 gating network $g^{(2)}$ selects the expert corresponding to the region $U_i$ such that $\bx \in U_i$.

By Theorem~\ref{thm: C^k approximation rate}, there exists $E$ two-layer networks $\tau_i$ ($i\in[E]$), with width $m_i\geq\Omega(E^2)$, such that:
\begin{align*}
    \norm{\rho_i-\tau_i}\leq\frac{1}{4E},\quad\forall i\in[E].
\end{align*}
Let $\tau$ be the concatenated two-layer network of the $E$ two-layer networks, i.e., $\tau(\bx)=(\tau_1(\bx),\cdots,\tau_E(\bx))^\top$, then it satisfies:
\begin{equation}\label{proof: thm: 1-layer, general: score error}
    \norm{(\rho_1,\cdots,\rho_E)^\top-\tau}\leq\frac{1}{4E}.
\end{equation}

In Layer 1, each expert network $f^{(1,i)}:\bbR^D\to\bbR^{D+E}$ shares the same form: 
\begin{align*}
    f^{(1,i)}(\bx)=\begin{pmatrix}
        \bx \\ \tau(\bx)
    \end{pmatrix}.
\end{align*}

The weight matrix $\bW_R^{(1)}$ of the gating network $g^{(1)}:\bbR^D\to\bbR^E$ can be any constant, such as $\bzero$.
In fact, Layer 1 behaves as a standard dense network computing $\tau$.

For the gating network $g^{(2)}$ in Layer 2, we define its weight matrix as:
\begin{align*}
   \bW_R^{(2)}=\big(\bzero_{E\times D},\bI_{E\times E}\big)\in\bbR^{E\times (D+E)}.
\end{align*}

Now we \underline{check the routing mechanism}: given an input $\bx\in\cM$, Layer 1 outputs 
\begin{align*}
    h^{(1)}(\bx)=\begin{pmatrix}
    \bx \\ \tau(\bx)
\end{pmatrix}\in\bbR^{D+E}.
\end{align*} 
Then the gating output of the Layer 2 is 
\begin{align*}
    g^{(2)}(h^{(1)}(\bx))=\bW_R^{(2)} h^{(1)}(\bx)=\tau(\bx)\in\bbR^E.
\end{align*}
Then it will assign $\bx$ to the $i_{\bx}$-th expert network in Layer 2, where
\begin{align*}
    i_{\bx}:=\argmax\limits_{i\in[E]} g_i^{(2)}(\bx)=\argmax\limits_{i\in[E]} \tau_i(\bx).
\end{align*}

Since $\sum_{i\in[E]}\rho_i(\bx)=1$ and $\rho_i\geq0,\forall i\in[E]$, we must have
$\max_{i\in[E]}\rho_i(\bx)\geq\frac{1}{E}$. Using~\eqref{proof: thm: 1-layer, general: score error}, we further have:
\begin{align*}
    \tau_{i_{\bx}}(\bx)=\max_{i\in[E]}\tau_i(\bx)\geq\max_{i\in[E]}\rho_i(\bx)-\frac{1}{4E}=\frac{3}{4E},
\end{align*}
which implies
\begin{align*}
    \rho_{i_{\bx}}(\bx)\geq \tau_{i_{\bx}}(\bx)-\frac{1}{4E}\geq\frac{3}{4E}-\frac{1}{4E}=\frac{1}{2E}>0.
\end{align*}

Thus,
\begin{equation}\label{proof: thm: 1-layer, general: exact assignment}
    \bx\in U_{i_{\bx}}.
\end{equation}

This establishes exact assignment of $\bx$ to its correct region.

\underline{\bf Expert networks (Layer 2).}

{\em Proof sketch:}
Each expert network $f^{(2,i)}$ in Layer 2 is responsible for {\em approximating the $d$-dimensional local target functions $f|_{U_i} \circ \phi_i^{-1}$ and the smooth chart maps $\phi_i$}.

By Theorem~\ref{thm: C^k approximation rate}, there exist two-layer dense networks $g_i$ and $\psi_i$, with width $m$, such that
\begin{align*}
    \norm{f|_{U_i}\circ\phi_i^{-1}-g_i}&\leq\tilde{\cO}\left(\max\left\{m^{-\frac{\kappa(f|_{U_i}\circ\phi_i^{-1})}{d}}, m^{-1/2}\right\}\right)
    \leq\tilde{\cO}\left(m^{-\frac{\kappa(f|_{U_i}\circ\phi_i^{-1})}{d}\wedge\frac{1}{2}}\right);
    \\
    \norm{\phi_i-\psi_i}&\leq\tilde{\cO}\left(m^{-1/2}\right).
\end{align*}

Let the expert network $f^{(2,i)}:\bbR^{D+E}\to\bbR$ as
\begin{align*}
    f^{(2,i)}(h^{(1)}(\bx)):=g_i\circ \psi_i(h^{(1)}_{1:E}(\bx))=g_i\circ \psi_i(\bx),
\end{align*} 

which is a three-layer dense network and satisfies:
\begin{equation}\label{proof: thm: 1-layer, general: expert approximation}
\begin{aligned}
    &\sup_{\bx\in U_i}\left|f|_{U_i}(\bx)-f^{(2,i)}(h^{(1)}(\bx))\right|
    =\sup_{\bx}\left|f|_{U_i}\circ\phi_i^{-1}\circ\phi_i(\bx)-g_i\circ \psi_i(\bx)\right|
    \\=&
    \norm{f|_{U_i}\circ\phi_i^{-1}\circ\phi_i-f|_{U_i}\circ\phi_i^{-1}\circ \psi_i+f|_{U_i}\circ\phi_i^{-1}\circ \psi_i-g_i\circ \psi_i}
    \\\leq&
     \norm{f|_{U_i}\circ\phi_i^{-1}\circ\phi_i-f|_{U_i}\circ\phi_i^{-1}\circ \psi_i}+\norm{f|_{U_i}\circ\phi_i^{-1}\circ \psi_i-g_i\circ \psi_i}
     \\\leq&
     \norm{f|_{U_i}\circ\phi_i^{-1}}\norm{\phi_i-\psi_i}+
     \norm{f|_{U_i}\circ\phi_i^{-1}-g_i}
     \\\leq&
     \tilde{\cO}\left(m^{-\frac{\kappa(f|_{U_i}\circ\phi_i^{-1})}{d}\wedge\frac{1}{2}}\right)=\tilde{\cO}\left(m^{-\frac{\kappa(f|_{U_i})}{d}\wedge\frac{1}{2}}\right).
\end{aligned}
\end{equation}

\underline{\bf The final estimate.}

Given an input $\bx\in\cM$, the routing mechanism assigns it to the correct expert network $f^{(2,i_{\bx})}$, which satisfies $\bx\in U_{i_{\bx}}$~\eqref{proof: thm: 1-layer, general: exact assignment}. Consequently, the approximation error holds:
\begin{align*}
    &|f(\bx)-f^{(2,i_{\bx})}(h^{(1)}(\bx))|
    =\Big|f|_{U_{i_{\bx}}}(\bx)-f^{(2,i_{\bx})}(h^{(1)}(\bx))\Big|
    \\\overset{\eqref{proof: thm: 1-layer, general: expert approximation}}{\leq}&\tilde{\cO}\left(m^{-\frac{\kappa(f|_{U_{i_{\bx}}})}{d}\wedge\frac{1}{2}}\right)
    \leq\max_{i\in[E]}\tilde{\cO}\left(m^{-\frac{\kappa(f|_{U_i})}{d}\wedge\frac{1}{2}}\right).
\end{align*}
Since $\bx$ is arbitrary, this concludes the proof.

Additionally, although this analysis is based on ReLU-activated experts, the theoretical framework extends naturally to MoEs with other activation functions (e.g., GeLU, SiLU, Swish) by using the results in~\citet{zhang2024deep}.

\subsection{Proof of Corollary~\ref{thm: 1-layer, special}}

The proof follows exactly the same structure as Theorem~\ref{thm: 1-layer, general}, except for a simplification in the step concerning the {\bf Expert networks (Layer 2)}. We illustrate only the modified part below.

\underline{\bf Expert networks (Layer 2).}

{\em Compared to Section~\ref{subappendix: proof shallow, main}, the key simplification is that the chart maps $\phi_i$ are \emph{linear} (see Example~\ref{example: manifold without boundary}). Thus, we do not need to approximate them with neural networks. We only need to approximate the composition $f|_{U_i} \circ \phi_i^{-1}$ using a two-layer network.}

By Theorem~\ref{thm: C^k approximation rate}, there exists a two-layer dense networks $g_i$, with width $m$, such that
\begin{align*}
    \norm{f|_{U_i}\circ\phi_i^{-1}-g_i}&\leq\tilde{\cO}\left(\max\left\{m^{-\frac{\kappa(f|_{U_i}\circ\phi_i^{-1})}{d}}, m^{-1/2}\right\}\right)
    \leq\tilde{\cO}\left(m^{-\frac{\kappa(f|_{U_i}\circ\phi_i^{-1})}{d}\wedge\frac{1}{2}}\right).
\end{align*}

Let the expert network $f^{(2,i)}:\bbR^{D+E}\to\bbR$ as
\begin{align*}
    f^{(2,i)}(h^{(1)}(\bx)):=g_i\circ \phi_i(h_{1:E}^{(1)}(\bx))=g_i\circ \phi_i(\bx),
\end{align*} 

which is a two-layer dense network since $\phi_i$ is linear and $g_i$ is two-layer. It satisfies:
\begin{equation}
\begin{aligned}
    &\sup_{\bx\in U_i}\left|f|_{U_i}(\bx)-f^{(2,i)}(h^{(1)}(\bx))\right|
    =\sup_{\bx}\left|f|_{U_i}\circ\phi_i^{-1}\circ\phi_i(\bx)-g_i\circ \phi_i(\bx)\right|
    \\=&
    \norm{f|_{U_i}\circ\phi_i^{-1}\circ \phi_i-g_i\circ \phi_i}
    \leq\norm{f|_{U_i}\circ\phi_i^{-1}-g_i}
     \\\leq&
     \tilde{\cO}\left(m^{-\frac{\kappa(f|_{U_i}\circ\phi_i^{-1})}{d}\wedge\frac{1}{2}}\right)=\tilde{\cO}\left(m^{-\frac{\kappa(f|_{U_i})}{d}\wedge\frac{1}{2}}\right).
\end{aligned}
\end{equation}

\vspace{1.cm}

\section{Proofs in Section~\ref{section: deep MoE}}
\label{appendix: proofs for deep MoE}

\subsection{Proof of Theorem~\ref{thm: deep MoE, general}}
\label{subappendix: proof deep, main}

The key idea is that for each $l\in[L]$, Layers $2l-1$ and $2l$ jointly approximate the subfunction 
\begin{align*}
    f_{l,i_l}(\bx_l), \text{ where } i_l\in\{j\in[E]:\bx_l\in U_{l,j}\}.
\end{align*}
The overall network composition then constructs the full function $f$ from these subcomponents, as defined in Equation~\eqref{equ: piecewise function, general}.

\underline{\bf Embedding.}
To facilitate layerwise operations, we first embed the input $\bx=(\bx_1^\top,\cdots,\bx_L^\top)^\top\in\bbR^{LD}$ into an extended space: 
\begin{align*}
    \bx^{(0)}=(\bx_1^\top,\cdots,\bx_L^\top,\bzero_{L}^\top)^\top\in\bbR^{LD+L},
\end{align*}
where the final $L$ entries are used to store the subfunctions approximated in subsequent layers. And we denote
\begin{align*}
    \tilde{D}:=LD+L.
\end{align*}

For simplicity, we denote the output of $2l$-th layer ($l\in[L]$) be $\bx^{(2l)}\in\bbR^{\tilde{D}}$.

Each two-layer MoE block operates similarly to the shallow case studied in Section~\ref{subappendix: proof shallow, main}, and thus the core proof strategy carries over. For completeness and clarity, we present the full proof below.

\underline{\bf $l$-th routing mechanism (Layer $2l-1$ and gating in Layer $2l$).}

{\em Proof sketch:} These components work together to {\em exactly assign $\bx_l$ to its correct expert $f^{(2l,i)}$}. Specifically, the first MoE layer $h^{(2l-1)}$ behaves like a dense model, approximating the smooth partition functions $\rho_{l,i}$'s. Then the Layer-2 gating network $g^{(2l)}$ selects the expert corresponding to the region $U_i$ such that $\bx_l \in U_{l,i}$.

By Theorem~\ref{thm: C^k approximation rate}, there exists $E$ two-layer networks $\tau_{l,i}$ ($i\in[E]$), with width $m_{l,i}\geq\Omega(E^2)$, such that:
\begin{align*}
    \norm{\rho_{l,i}-\tau_{l,i}}\leq\frac{1}{4E},\quad\forall i\in[E].
\end{align*}
Let $\tau^{(l)}$ be the concatenated two-layer network of the $E$ two-layer networks, i.e., $\tau^{(l)}(\bx_l)=(\tau_{l,i}(\bx_l),\cdots,\tau_{l,E}(\bx_l))^\top$, then it satisfies:
\begin{equation}\label{proof: thm: deep, general: score error}
    \norm{(\rho_{l,1},\cdots,\rho_{l,E})^\top-\tau^{(l)}}\leq\frac{1}{4E}.
\end{equation}

In Layer $2l-1$, each expert network $f^{(2l-1,i)}:\bbR^{\tilde{D}}\to\bbR^{\tilde{D}+E}$ shares the same form: 
\begin{align*}
    f^{(2l-1,i)}(\bx^{(2l-2)})=\begin{pmatrix}
        \bx^{(2l-2)} \\ \tau^{(l)}(\bx_l)
    \end{pmatrix}.
\end{align*}

The weight matrix $\bW_R^{(2l-1)}$ of the gating network $g^{(2l-1)}:\bbR^{\tilde{D}}\to\bbR^E$ can be any constant, such as $\bzero$.
In fact, Layer $2l-1$ behaves as a standard dense network computing $\tau^{(l)}(\bx_l)$.

For the gating network $g^{(2l)}$ in Layer $2l$, we define its weight matrix as:
\begin{align*}
   \bW_R^{(2l)}=\big(\bzero_{E\times \tilde{D}},\bI_{E\times E}\big)\in\bbR^{E\times (\tilde{D}+E)}.
\end{align*}

Now we \underline{check the routing mechanism}: given an input $\bx=(\bx_1,\cdots,\bx_L)\in\cM$, where $\bx_l\in\cM_l$, Layer $2l-1$ outputs 
\begin{align*}
    \bx^{(2l-1)}=h^{(2l-1)}(\bx^{(2l-2)})=\begin{pmatrix}
    \bx^{(2l-2)} \\ \tau^{(l)}(\bx_l)
\end{pmatrix}\in\bbR^{\tilde{D}+E}.
\end{align*}
Then the gating output of the Layer $2l$ is 
\begin{align*}
    g^{(2l)}(\bx^{(2l-1)})=\bW_R^{(2l)} h^{(2l-1)}(\bx_l)=\tau^{(l)}(\bx_l)\in\bbR^E.
\end{align*}
Then it will assign $\bx_l$ to the $i_{\bx_l}$-th expert network in Layer 2, where
\begin{align*}
    i_{\bx_l}:=\argmax\limits_{i\in[E]} g_i^{(2l)}(\bx_l)=\argmax\limits_{i\in[E]} \tau_{l,i}(\bx_l).
\end{align*}

Since $\sum_{i\in[E]}\rho_{l,i}(\bx_l)=1$ and $\rho_{l,i}\geq0,\forall i\in[E]$, we must have
$\max_{i\in[E]}\rho_{l,i}(\bx_l)\geq\frac{1}{E}$. Using~\eqref{proof: thm: deep, general: score error}, we further have:
\begin{align*}
    \tau_{i_{\bx_l}}(\bx_l)=\max_{i\in[E]}\tau_{l,i}(\bx_l)\geq\max_{i\in[E]}\rho_{l,i}(\bx_l)-\frac{1}{4E}=\frac{3}{4E},
\end{align*}
which implies
\begin{align*}
    \rho_{i_{\bx_l}}(\bx_l)\geq \tau_{i_{\bx_l}}(\bx_l)-\frac{1}{4E}\geq\frac{3}{4E}-\frac{1}{4E}=\frac{1}{2E}>0.
\end{align*}

Thus,
\begin{equation}\label{proof: thm: deep, general: exact assignment}
    \bx_l\in U_{i_{\bx_l}}.
\end{equation}

This establishes exact assignment of $\bx_l$ to its correct region.

\underline{\bf $l$-th expert networks (Layer $2l$).}

{\em Proof sketch:}
Each expert network $f^{(2l,i)}$ in Layer 2 is responsible for {\em approximating the $d$-dimensional local target functions $f_{l,i} \circ \phi_{l,i}^{-1}$ and the smooth chart maps $\phi_{l,i}$}.

By Theorem~\ref{thm: C^k approximation rate}, there exist two-layer dense networks $g_{l,i}$ and $\psi_{l,i}$, with width $m$, such that
\begin{align*}
    \norm{f_{l,i}\circ\phi_{l,i}^{-1}-g_{l,i}}&\leq\tilde{\cO}\left(\max\left\{m^{-\frac{\kappa(f_{l,i}\circ\phi_{l,i}^{-1})}{d_l}}, m^{-1/2}\right\}\right)
    \leq\tilde{\cO}\left(m^{-\frac{\kappa(f_{l,i}\circ\phi_{l,i}^{-1})}{d_l}\wedge\frac{1}{2}}\right);
    \\
    \norm{\phi_{l,i}-\psi_{l,i}}&\leq\tilde{\cO}\left(m^{-1/2}\right).
\end{align*}

Let the expert network $f^{(2l,i)}:\bbR^{\tilde{D}+E}\to\bbR^{\tilde{D}}$ as
\begin{align*}
    f^{(2l,i)}(\bx^{(2l-1)}):=\begin{pmatrix}
        \bx \\ \bzero_{l-1} \\
        g_{l,i}\circ \psi_{l,i}(\bx_{1+(l-1)E:\ lE}^{(2l-1)}) \\ \bzero_{E-l}
    \end{pmatrix}
    =\begin{pmatrix}
        \bx \\ \bzero_{l-1} \\
        g_{l,i}\circ \psi_{l,i}(\bx_{l}) \\ \bzero_{E-l}
    \end{pmatrix},
\end{align*}

which is a three-layer dense network and satisfies:
\begin{equation}\label{proof: thm: deep, general: expert approximation}
\begin{aligned}
    &\sup_{\bx_l\in U_{l,i}}\left|f_{l,i}(\bx_l)-f_{\tilde{D}+l}^{(2l,i)}(\bx^{(2l-1)})\right|
    =\sup_{\bx_l\in U_{l,i}}\left|f_{l,i}\circ\phi_{l,i}^{-1}\circ\phi_{l,i}(\bx_l)-g_{l,i}\circ \psi_{l,i}(\bx_l)\right|
    \\=&
    \norm{f_{l,i}\circ\phi_{l,i}^{-1}\circ\phi_{l,i}-f_{l,i}\circ\phi_{l,i}^{-1}\circ \psi_{l,i}+f_{l,i}\circ\phi_{l,i}^{-1}\circ \psi_{l,i}-g_{l,i}\circ \psi_{l,i}}
    \\\leq&
     \norm{f_{l,i}\circ\phi_{l,i}^{-1}\circ\phi_{l,i}-f_{l,i}\circ\phi_{l,i}^{-1}\circ \psi_{l,i}}+\norm{f_{l,i}\circ\phi_{l,i}^{-1}\circ \psi_{l,i}-g_{l,i}\circ \psi_{l,i}}
     \\\leq&
     \norm{f_{l,i}\circ\phi_{l,i}^{-1}}\norm{\phi_{l,i}-\psi_{l,i}}+
     \norm{f_{l,i}\circ\phi_{l,i}^{-1}-g_{l,i}}
     \\\leq&
     \tilde{\cO}\left(m^{-\frac{\kappa(f_{l,i}\circ\phi_{l,i}^{-1})}{d_l}\wedge\frac{1}{2}}\right)=\tilde{\cO}\left(m^{-\frac{\kappa(f_{l,i})}{d_l}\wedge\frac{1}{2}}\right).
\end{aligned}
\end{equation}

\underline{\bf The final estimate.}

From the above construction, the output of the $2L$-th layer takes the form:
\begin{align*}
    \bx^{(2L)}=\begin{pmatrix}
        \bx \\
        g_{1,i}\circ\psi_{1,i_{\bx_1}}(\bx_1) \\ \vdots \\
        g_{L,i}\circ\psi_{L,i_{\bx_L}}(\bx_L)
    \end{pmatrix}\in\bbR^{LD+L}.
\end{align*}
We decode this vector by applying a linear projection that extracts the last $L$ components, yielding:
\begin{align*}
    \Psi(\bx):=\begin{pmatrix}
        \bzero_{L\times LD}, \bI_{L\times L}
    \end{pmatrix} \bx^{(2L)}
    =\begin{pmatrix}
        g_{1,i}\circ\psi_{1,i_{\bx_1}}(\bx_1) \\ \vdots \\
        g_{L,i}\circ\psi_{L,i_{\bx_L}}(\bx_L)
    \end{pmatrix}\in\bbR^{L}.
\end{align*}

Now consider an input $\bx=(\bx_1,\cdots,\bx_L)\in\cM$.
For each $l\in[L]$, the $l$-th routing mechanism assigns it to the correct expert network $f^{(2l,i_{\bx_l})}$, which satisfies $\bx_l\in U_{i_{\bx_l}}$~\eqref{proof: thm: deep, general: exact assignment}. Consequently, the approximation error holds:
\begin{align*}
    &\|f(\bx)-\Psi(\bx)\|
    =\max_{l\in[L]}\Big|f_{l,i_{\bx_l}}(\bx_l)-g_{l,i_{\bx_l}}\circ\psi_{l,i_{\bx_l}}(\bx_l)\Big|
    \\\overset{\eqref{proof: thm: deep, general: expert approximation}}{\leq}&\max_{l\in[L]}\tilde{\cO}\left(m^{-\frac{\kappa(f_{l,i_{\bx_l}})}{d_l}\wedge\frac{1}{2}}\right)
    \leq\max_{l\in[L]}\max_{i\in[E]}\tilde{\cO}\left(m^{-\frac{\kappa(f_{l,i})}{d_l}\wedge\frac{1}{2}}\right).
\end{align*}
Since $\bx$ is arbitrary, this concludes the proof.

\subsection{Proof of Theorem~\ref{thm: deep MoE, warmup}}

The proof follows the same structure as Theorem~\ref{thm: deep MoE, general}, except for a simplification in the construction of the {\bf routing mechanism}. For completeness and clarity, we provide the full proof below.

The key idea is that for each $l\in[L]$, Layers $2l-1$ and $2l$ jointly approximate the subfunction 
\begin{align*}
    f_{l,i_l}(x_l), \text{ where } i_l\in\{j\in[E]:x_l\in U_{l,j}=[j-1,j]\}.
\end{align*}
The overall network composition then constructs the full function $f$ from these subcomponents.

For simplicity, we denote the output of $2l$-th layer ($l\in[L]$) be $\bx^{(2l)}\in\bbR^{L}$.

\underline{\bf $l$-th routing mechanism (Layer $2l-1$ and gating in Layer $2l$).}

{\em Proof sketch:} These components work together to {\em exactly assign $\bx_l$ to its correct expert $f^{(2l,i)}$}. Specifically, the first MoE layer $h^{(2l-1)}$ behaves like a dense model, serving as an indicator function. 
Then the Layer-2 gating network $g^{(2l)}$ selects the expert corresponding to the region $U_i$ such that $x_l \in U_{l,i}$.

Consider the specific indicator function comprising 3 ReLU neurons:
\begin{align*}
   \tau_{l,i}(x_l):= {\rm ReLU}(x_l-(i-1))+{\rm ReLU}(x_l-i)-2{\rm ReLU}(x_l-(i-1/2)),\quad i\in[E].
\end{align*}

which satisfies:
\begin{equation}\label{proof: thm: deep, 1-dim: ReLU property}
    \tau_{l,i}(x_l)\begin{cases}
   >0\quad \text{if } x_l\in(i-1,i)\\
   0\quad\quad\text{else}
   \end{cases}.
\end{equation}

Let $\tau^{(l)}$ be the concatenated two-layer network of the $E$ two-layer networks, i.e., $\tau^{(l)}(\bx_l)=(\tau_{l,i}(\bx_l),\cdots,\tau_{l,E}(\bx_l))^\top$. Note that the width of $\tau^{(l)}$ is only $3E$.

In Layer $2l-1$, each expert network $f^{(2l-1,i)}:\bbR^{L}\to\bbR^{L+E}$ shares the same form: 
\begin{align*}
    f^{(2l-1,i)}(\bx^{(2l-2)})=\begin{pmatrix}
        \bx^{(2l-2)} \\ \tau^{(l)}(x_l)
    \end{pmatrix}.
\end{align*}

The weight matrix $\bW_R^{(2l-1)}$ of the gating network $g^{(2l-1)}:\bbR^L\to\bbR^E$ can be any constant, such as $\bzero$.
In fact, Layer $2l-1$ behaves as a standard dense network computing $\tau^{(l)}(x_l)$.

For the gating network $g^{(2l)}$ in Layer $2l$, we define its weight matrix as:
\begin{align*}
   \bW_R^{(2l)}=\big(\bzero_{E\times L},\bI_{E\times E}\big)\in\bbR^{E\times (L+E)}.
\end{align*}

Now we \underline{check the routing mechanism}: given an input $\bx=(x_1,\cdots,x_L)\in[0,E]^L$, where $x_l\in[0,E]$, Layer $2l-1$ outputs 
\begin{align*}
    \bx^{(2l-1)}=h^{(2l-1)}(\bx^{(2l-2)})=\begin{pmatrix}
    \bx^{(2l-2)} \\ \tau^{(l)}(x_l)
\end{pmatrix}\in\bbR^{L+E}.
\end{align*}
Then the gating output of the Layer $2l$ is 
\begin{align*}
    g^{(2l)}(\bx^{(2l-1)})=\bW_R^{(2l)} h^{(2l-1)}(x_l)=\tau^{(l)}(x_l)\in\bbR^E.
\end{align*}
Then it will assign $x_l$ to the $i_{x_l}$-th expert network in Layer 2, where
\begin{align*}
    i_{x_l}:=\argmax\limits_{i\in[E]} g_i^{(2l)}(x_l)=\argmax\limits_{i\in[E]} \tau_{l,i}(x_l).
\end{align*}

From the property~\eqref{proof: thm: deep, 1-dim: ReLU property}, we have:
\begin{equation}\label{proof: thm: deep, 1-dim: exact assignment}
    x_l\in U_{i_{x_l}}.
\end{equation}

This establishes exact assignment of $x_l$ to its correct region.

\underline{\bf $l$-th expert networks (Layer $2l$).}

{\em Proof sketch:}
Each expert network $f^{(2l,i)}$ in Layer 2 is responsible for {\em approximating the $1$-dimensional local target functions $f_{l,i}$}.

By Theorem~\ref{thm: C^k approximation rate}, there exist two-layer dense networks $g_{l,i}$, with width $m$, such that
\begin{align*}
    \norm{f_{l,i}-g_{l,i}}&\leq\tilde{\cO}\left(\max\left\{m^{-\kappa(f_{l,i})}, m^{-1/2}\right\}\right)
    \leq\tilde{\cO}\left(m^{-\kappa(f_{l,i})\wedge\frac{1}{2}}\right).
\end{align*}

Let the expert network $f^{(2l,i)}:\bbR^{L+E}\to\bbR^{L}$ as
\begin{align*}
    f^{(2l,i)}(\bx^{(2l-1)})
    =\begin{pmatrix}
        \bx_{1:l-1}^{(2l-1)}\\
        g_{l,i}(x_{l}) \\ \bx_{l+1:E}^{(2l-1)}
    \end{pmatrix},
\end{align*}

which is a three-layer dense network and satisfies:
\begin{equation}\label{proof: thm: deep, 1-dim: expert approximation}
\begin{aligned}
    &\sup_{x_l\in U_{l,i}}\left|f_{l,i}(x_l)-f_{l}^{(2l,i)}(\bx^{(2l-1)})\right|
    =\sup_{x_l\in U_{l,i}}\left|f_{l,i}(x_l)-g_{l,i}(x_l)\right|
     \\\leq&
     \norm{f_{l,i}-g_{l,i}}
     \leq
     \tilde{\cO}\left(m^{-\kappa(f_{l,i}\circ\phi_{l,i}^{-1})\wedge\frac{1}{2}}\right)=\tilde{\cO}\left(m^{-\kappa(f_{l,i})\wedge\frac{1}{2}}\right).
\end{aligned}
\end{equation}

\underline{\bf The final estimate.}

From the above construction, the output of the $2L$-th layer takes the form:
\begin{align*}
    \Psi(\bx):=\bx^{(2L)}=\begin{pmatrix}
        g_{1,i_{x_1}}(x_1) \\ \vdots \\
        g_{L,i_{x_L}}(x_L)
    \end{pmatrix}\in\bbR^{L}.
\end{align*}

Now consider an input $\bx=(x_1,\cdots,x_L)\in[0,E]^L$.
For each $l\in[L]$, the $l$-th routing mechanism assigns it to the correct expert network $f^{(2l,i_{x_l})}$, which satisfies $x_l\in U_{i_{x_l}}$~\eqref{proof: thm: deep, 1-dim: exact assignment}. Consequently, the approximation error holds:
\begin{align*}
    &\|f(\bx)-\Psi(\bx)\|
    =\max_{l\in[L]}\Big|f_{l,i_{x_l}}(x_l)-g_{l,i_{x_l}}(x_l)\Big|
    \\\overset{\eqref{proof: thm: deep, 1-dim: expert approximation}}{\leq}&\max_{l\in[L]}\tilde{\cO}\left(m^{-\kappa(f_{l,i_{x_l}})\wedge\frac{1}{2}}\right)
    \leq\max_{l\in[L]}\max_{i\in[E]}\tilde{\cO}\left(m^{-\kappa(f_{l,i})\wedge\frac{1}{2}}\right).
\end{align*}
Since $\bx$ is arbitrary, this concludes the proof.

\vspace{1.cm}

\section{Experimental Details}
\label{appendix: experiments}

The experiments in Section~\ref{section: experiments} are conducted on 1 A100 GPU.

In Experiment I, the models are trained for $2,000$ iterations with batch size $128$ (online), using squared loss and Adam optimizer with learning rate \texttt{1e-3}.

In Experiment II, the models are trained for $5,000$ iterations with batch size $128$ (online), using squared loss and Adam optimizer with learning rate \texttt{1e-3}.

\end{document}